\documentclass{article}

     \PassOptionsToPackage{numbers, compress}{natbib}
  \usepackage[preprint]{neurips_2021_mod}
\usepackage[utf8]{inputenc} % allow utf-8 input
\usepackage[T1]{fontenc}    % use 8-bit T1 fonts
\usepackage{hyperref}       % hyperlinks
\usepackage{url}            % simple URL typesetting
\usepackage{booktabs}       % professional-quality tables
\usepackage{amsfonts}       % blackboard math symbols
\usepackage{nicefrac}       % compact symbols for 1/2, etc.
\usepackage{microtype}      % microtypography

\usepackage{multirow}
\usepackage{amsmath,amssymb,amsthm,amscd,amstext}
\usepackage{algorithm,algorithmic}
\usepackage{wrapfig}
\usepackage{capt-of}

\usepackage{bbm}
\usepackage{relsize}

\usepackage[toc,page,header]{appendix}
\usepackage{minitoc}

\newcommand{\gv}{{\boldsymbol g}}

\usepackage{soul}

\newcommand{\EE}{\mathbb{E}}

\newcommand{\bs}[1]{\boldsymbol{#1}}
\newcommand{\BR}{\mathbb{R}}

\newcommand{\CX}{\mathcal{X}}

\newcommand{\CZ}{\mathcal{Z}}
\newcommand{\MI}{\text{MI}}
\newcommand{\CE}{\mathcal{E}}
\newcommand{\CF}{\mathcal{F}}

\newcommand{\Enc}{\text{Enc}}

\newcommand{\CL}{\mathcal{L}}

\renewcommand{\CD}{\mathcal{D}}

\usepackage{tikz}
\theoremstyle{plain}
\newtheorem{thm}{Theorem}[section] % reset theorem numbering for each chapter
\theoremstyle{definition}
\newtheorem{defn}[thm]{Definition} % definition numbers are dependent on theorem numbers
\newtheorem{prop}[thm]{Proposition} % same for example numbers
\newtheorem{assumption}[thm]{Assumption}
\newtheorem{lem}[thm]{Lemma}

\newtheorem{col}[thm]{Corollary}

\newcommand{\beq}{\begin{equation}}
\newcommand{\eeq}{\end{equation}}
\newcommand{\beqs}{\begin{eqnarray}}
\newcommand{\eeqs}{\end{eqnarray}}
\newcommand{\barr}{\begin{array}}
\newcommand{\earr}{\end{array}}

\newcommand{\ESS}{\text{ESS}}

\newcommand{\infonce}{\texttt{InfoNCE}}

\newcommand{\BA}{\texttt{BA}}

\newcommand{\NWJ}{\texttt{NWJ}}
\newcommand{\DV}{\texttt{DV}}
\newcommand{\MINE}{\texttt{MINE}}

\newcommand{\FLO}{\texttt{FLO}}

\newcommand{\FLAT}{\texttt{FlatNCE}}

\newcommand{\SimCLR}{\texttt{SimCLR}}
\newcommand{\SimCSE}{\texttt{SimCSE}}
\newcommand{\CLIP}{\texttt{CLIP}}
\newcommand{\ALIGN}{\texttt{ALIGN}}
\newcommand{\FCLR}{\texttt{FlatCLR}}

\newcommand{\CT}{\mathcal{T}}

\title{Simpler, Faster, Stronger: Breaking The $\log$-$K$ Curse On Contrastive Learners With FlatNCE}
\author{Junya Chen${}^{1,\dagger}$, Zhe Gan${}^2$, Xuan Li${}^3$, Qing Guo${}^3$, Liqun Chen${}^4$, Shuyang Gao${}^4$, Wenlian Lu${}^5$\\
{\bf Tagyoung Chung${}^4$, Yi Xu${}^4$, Belinda Zeng${}^4$, Fan Li${}^1$, Lawrence Carin${}^6$, Chenyang Tao${}^{1,\dagger}$} \\ ${}^1$Duke University  ${}^2$Microsoft ${}^3$Virginia Tech ${}^4$Amazon ${}^5$Fudan University ${}^6$KAUST}

\begin{document}

\maketitle

%%%%% To create table of contents only for Appendix %%%%%%

\doparttoc % Tell to minitoc to generate a toc for the parts
\faketableofcontents % Run a fake tableofcontents command for the partocs

%\part{} % Start the document part
%\parttoc % Insert the document TOC

%%%%%%%%%%%%%%%%%%%%%%%%%%%%%%%%%%

\begin{abstract}
%a significant drawback of 
%Dealing with severe class imbalance poses a major challenge for real-world applications, especially when the accurate classification and generalization of minority classes is of primary interest ({\it e.g.}, insurance fraud, severe weather, traffic safety, rare disease, etc.). While existing solutions mostly focus on sampling or weighting adjustments to alleviate the pathological imbalance, or imposing inductive bias to prioritize non-spurious associations, we take novel perspectives to promote sample efficiency and model generalization based on causal invariance. Our proposal posits a meta-distributional scenario, where the data generating mechanism is invariant across the label-conditional feature distributions. Such transfer assumption enables efficient knowledge transfer from the dominant classes to their under-represented counterparts, even if the respective feature distributions are apparently different. A principled data inflation procedure is devised to enlarge the representation of minority classes. Our development is orthogonal to existing extreme classification techniques and can be seamlessly integrated. The utility of our proposal is validated with an extensive set of synthetic and real-world benchmarks. 
$\infonce$-based contrastive representation learners, such as $\SimCLR$~\cite{chen2020simple}, have been tremendously successful in recent years. However, these contrastive schemes are notoriously resource demanding, as their effectiveness breaks down with small-batch training ({\it i.e.}, the $\log$-$K$ curse, whereas $K$ is the batch-size). 
In this work, we reveal mathematically why contrastive learners fail in the small-batch-size regime, and present a novel simple, non-trivial contrastive objective named $\FLAT$, which fixes this issue. Unlike $\infonce$, our $\FLAT$ no longer explicitly appeals to a discriminative classification goal for contrastive learning. Theoretically,  we show $\FLAT$ is the mathematical dual formulation of $\infonce$, thus bridging the classical literature on energy modeling; and empirically, we demonstrate that, with minimal modification of code, $\FLAT$ enables immediate performance boost independent of the subject-matter engineering efforts. The significance of this work is furthered by the powerful generalization of contrastive learning techniques, and the introduction of new tools to monitor and diagnose contrastive training. We substantiate our claims with empirical evidence on CIFAR10 and ImageNet datasets, where $\FLAT$ consistently outperforms $\infonce$.
\end{abstract}

%\vspace{-3pt}
\section{Introduction}
\vspace{-5pt}

%{\it Contrastive representation learning} has gained considerable momentum in recent years, largely credited to both its astonishing effectiveness in self-supervised learning setups \citep{tishby2015deep, hjelm2019learning} and fool-proof easy implementation \citep{oord2018representation}. 
As a consequence of the superior effectiveness in self-supervised learning setups \citep{tishby2015deep, hjelm2019learning} and their relatively easy implementation \citep{oord2018representation}, {\it contrastive representation learning} has gained considerable momentum in recent years. Successful applications have been reported in computer vision \citep{wu2018unsupervised, he2020momentum, chen2020simple, misra2020self}, natural language processing \citep{gao2021simcse, radford2021learning}, reinforcement learning \citep{laskin2020curl}, fairness \citep{gupta2021controllable}, amongst many others. However, contrastive learners are often resource demanding, as their effectiveness breaks down with small-batch training \citep{gutmann2012noise, poole2019variational}. This limits potential applications for very complex models or budgeted applications. In this study, we revisit the mathematics of contrastive learning to not only find practical remedies, but also suggest new research directions. 
%often found to be ineffective in the small batch-size regime. This limits their applications for very complex models or budgeted applications. In this study, we revisit the maths of contrastive learning to seek effective remedies. 

Originally developed for nonparametric density estimation, the idea of learning by contrasting {\it positive} and {\it negative} samples has deep roots in statistical modeling \citep{hinton2002training}. In the seminal work of \citep{gutmann2012noise}, its connection to discriminative classification was first revealed, and early utilization of the idea was celebrated by the notable success in training word embeddings \citep{mikolov2013distributed, mnih2013learning}. Framed under the name {\it negative sampling} \citep{ma2018noise}, contrastive techniques have been established as indispensable tools in scaling up the learning of intractable statistical models such as graphs \citep{perozzi2014deepwalk, grover2016node2vec}. 
%, with prominent applications as contrastive divergence optimization for {\it restricted Boltzmann machines}
%It quickly led to successful the application 
%in the learning of intractable statistical models such as graphs \citep{}.
%It quickly led to successful the application of distributed learning of word-embedding. 
%, with its recent adoption in representation learning proved tremendously fruitful. 

%Poineering applications primarily focused on the nonparametric estimation of data densities. 

%More recently, there has been renewed interest in .
More recently, surging interest in contrastive learners was sparked by the renewed understanding that connects to {\it mutual information} estimation \citep{oord2018representation, poole2019variational}. Fueled by the discovery of efficient algorithms and strong performance \citep{chen2020simple}, extensive research has been devoted to this active topic \citep{le2020contrastive}. These efforts range from theoretical investigations such as generalization error analyses \citep{arora2019theoretical} and asymptotic characterizations \citep{wang2020understanding}, to more practical aspects including hard-negative reinforcement \citep{robinson2020contrastive, kalantidis2020hard}, and sampling bias adjustment \citep{chuang2020debiased}. Along with various subject matter improvements \citep{logeswaran2018efficient, ozair2019wasserstein, henaff2020data, wu2020importance, gao2021simcse, radford2021learning}, contrastive learners now provide comprehensive solutions for self-supervised learning. 
%contrastive learning 
%that massively expedite computations

Despite encouraging progress, there are still many unresolved issues with contrastive learning, with the following three  particularly relevant to this investigation: ($i$) contrastive learners need a very large number of negative samples to work well; ($ii$) the bias, variance, and performance tradeoffs are in debate \citep{poole2019variational, arora2019theoretical, nozawa2021understanding}; and, crucially, ($iii$) there is a lack of training diagnostic tools for contrastive learning. Among these three, ($i$) is most concerning, as it implies training can be very expensive, while  
%Table \ref{tab:cost} \zhe{where it table 2?}\cytao{Currently at the end of paper, feels more appropriate in the Appendix.} summarizes the costs associated with representative contrastive models. Such
the needed massive computational resources may not be widely available. Even when such computational resources are accessible, the costs are prohibitive, and arguably entail a large carbon footprint.  
%\zhe{I feel we are over-selling, do we solve this?}\cytao{For Cifar10, yes. For ImageNet, I guess we just need a bit more time.}
%, to the extent that is nearly cost prohibitive

%\begin{figure}[t!]
%\begin{center}
%\includegraphics[height=2in, width=.7\textwidth]{figures/results/mi_opt}
%\end{center}
%\vspace{-1em}
%\caption{Comparison of empirical and ground-truth MI for $\infonce$ and $\FLAT$ on \texttt{Cifar}. Empirical MI is estimated with $K=128$ negative samples and used for training, while the ground-truth MI is obtained with $K=10k$ negative samples for evaluation. $\infonce$ representation creases to improve after its empirical MI reaches the $\log K$ cap, a limitation that $\FLAT$ overcomes. }
%\vspace{-.5em}
%\end{figure}

\begin{wrapfigure}[15]{R}{0.5\textwidth}
\vspace{-2.8em}
\scalebox{1.}{
%\hspace{-.5em}
\begin{minipage}{.45\textwidth}
\begin{figure}[H]
\begin{center}{
\hspace{-1.3em}\includegraphics[width=1.1\textwidth]{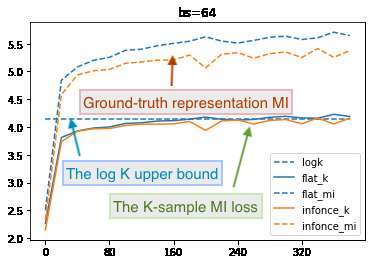}
}
\vspace{-2em}
% \caption{\small {$\FLAT$} better optimizes the true MI with a finite sample loss. \label{fig:mi_opt}}
\caption{\small $\FLAT$ continues to robustly optimize the ground-truth MI for representation even after the finite sample loss has saturated at $\log K$. \label{fig:mi_opt}}
%{$\FLAT$} learns a representation with better ground-truth MI even if its $K$-sample MI loss is tied with that of $\infonce$.
\end{center}
% updated policy sampler
\vspace{-1.5em}
\end{figure}
\end{minipage}
}
\end{wrapfigure}

We believe addressing ($ii$) and ($iii$) holds the key to resolving ($i$). A major inconsistency between theory and practice is that, contrary to expectation, more biased estimators such as $\infonce$ work better in practice than their tighter counterparts \citep{tschannen2020mutual}. The prevailing conjecture is that these biased contrastive learners benefit from a lower estimation variance \citep{arora2019theoretical, song2020understanding}. However, this conjecture is mostly based on experimental observations rather than formal variance analyses \citep{mcallester2018formal}, and the comparison is not technically fair since the less biased estimators use far less samples \citep{poole2019variational, bachman2019learning, hjelm2019learning}. Such incomplete understandings are partly caused by the absence of proper generic diagnostic tools to analyze contrastive learners. In this study, we hope to improve both the understanding and practice of contrastive representation learning via bridging these gaps.
%A major inconsistency between theoretical predictions and empirical observations is that, . 

%\begin{table}[t!]
%\begin{center}
%\caption{Cost of training a contrastive learner \label{tab:cost}}
%\begin{tabular}{cccc}
%\toprule
%Model & Sponsor & Neg. Size & Infrastructure\\
%\midrule
%\texttt{MoCo} \citep{he2020momentum} & Facebook & $65,536$ & $64$ V100 GPUs \\
%\texttt{SimCLR} \citep{chen2020simple} & Google &  $4,096$ & $128$-core TPU-v3\\
%\texttt{CLIP} \citep{radford2021learning} & OpenAI & $32,768$ & $592$ V100 GPUs\\
%\bottomrule
%\end{tabular}
%\end{center}
%\end{table}

Our development starts with two simple intuitions: ($i$) the contrasts between positive and negative data should be as large as possible, and ($ii$) the objective should be properly normalized to yield minimal variance. These heuristics lead to a simple, powerful, theoretically grounded novel contrastive learning objective we call $\FLAT$. We show $\FLAT$ is the mathematical dual of $\infonce$ \citep{poole2019variational}, a widely used MI estimator that empowers models such as $\SimCLR$ \citep{chen2020simple}, $\SimCSE$ \citep{gao2021simcse}, %\zhe{why even cite this paper? not proper here, why not moco?}\cytao{Reminiscent of our first draft. I am not aware of the CLIP \& ALIGN work and want to add some diversity in NLP. SimCSE is well-known in NLP I think.}
$\CLIP$ \citep{radford2021learning} and $\ALIGN$ \citep{jia2021scaling}. What makes $\FLAT$ unique is that it has deep roots in statistical physics, which enables it to optimize beyond the finite-sample bottleneck that has plagued $\infonce$ (Figure \ref{fig:mi_opt}). 
% \zhe{figure 1 is not clear.} \cytao{It's just a sketch and will be replaced with a proper remake. x is the training epoch and y is the mutual information. }

%Behind its simple ideology, $\FLAT$ is essentially a formal variational lower bound to the mutual information. What distinguishes $\FLAT$ from its predecessor $\infonce$ is that, $\FLAT$ targets a non-discriminative goal, and it is derived from an energy modeling perspective. 

Importantly, our research brings new insights into contrastive learning. While the new energy perspective continues to reinforce the heuristic of contrastive learning, $\FLAT$ shows appealing to a cross-entropy-based predictive objective is suboptimal. This echoes recent attempts in building non-discriminative contrastive learners \citep{wang2020understanding}, and to the best of our knowledge, we provide the first of its kind that comes with rigorous theoretical guarantees. Further, $\FLAT$ inspires a set of diagnostic tools that will benefit the contrastive learning community as a whole \citep{arora2019theoretical}.% We conclude this paper with an open invitation to solve some interesting challenges that we are actively exploring.

%While continuing to reinforce the heuristic that positive pairs should be distinct from the negative pairs, $\FLAT$ shows that appealing to the predictive goal implemented by cross-entropy is not absolutely unnecessary, and alternatives might even yield better results. Although some prior-arts have investigated non-CE alternatives for contrastive learning \citep{wang2020understanding}, to the best of our knowledge, we provide the first of its kind that comes with rigorous theoretical guarantees. Further, the formulation of $\FLAT$ inspires a set of diagnostic tools that will benefit contrastive learning community as a whole \citep{arora2019theoretical}. We conclude this paper with an open invitation to solve some interesting challenges that we are actively exploring.

\vspace{-5pt}
\section{Contrastive Representation Learning with InfoNCE}
\vspace{-8pt}

This section reviews the technical aspects of $\infonce$, how it is used to empower representation learning, and explains why $\infonce$ fails in small-batch training, hence motivating our work. 

%In this section, we briefly review the technical aspects of $\infonce$, and how it is used to empower representation learning. To motivate our subsequent development, we conclude this section with the insights on why $\infonce$ works suboptimally in the small batch-size regime. 

\vspace{-5pt}
\subsection{InfoMax with noise contrastive estimation}
\vspace{-3pt}

Estimation of {\it mutual information} (MI) is central to many scientific investigations and engineering problems \citep{shannon1948mathematical, mackay2003information, reshef2011detecting}, thus motivating a plethora of practical procedures \citep{gretton2003kernel, kraskov2004estimating, perez2008estimation, gao2015efficient, suzuki2008approximating, belghazi2018mutual, moyer2018invariant, cheng2020club}. Recently, burgeoning deep learning applications have raised particular interest in variational estimators of MI \citep{poole2019variational}, for they usually deliver stable performance and are easily amendable to gradient-based optimization. Among the many variational MI estimators, $\infonce$ stands out for its conceptual simplicity, ease of implementation and, crucially, strong empirical performance.% \citep{}.%\zhe{lack cite?} 

$\infonce$ is a multi-sample mutual information estimator built on the idea of {\it noise contrastive estimation} (NCE) \citep{gutmann2010noise}\footnote{In some contexts, it is also known as {\it negative sampling} \citep{ma2018noise}.}. It was first described in \citep{oord2018representation} under the name {\it contrastive predictive coding} (CPC), and later formalized and coined $\infonce$ in the work of \citep{poole2019variational}. Formally defined by\footnote{This is technically equivalent to the original definition due to the symmetry of $K$ samples.}
\beq
\label{eq:infonce}
I_{\infonce}^K(X;Y|f)\triangleq \EE_{p^K(x,y)}\left[\log \frac{f(x_1,y_1^{\oplus})}{\frac{1}{K}\sum_{k'} f(x_{1}, y_{k'}^{\ominus})}\right], I_{\infonce}^K(X;Y) \triangleq \underset{f\in \CF}{\max} \{I_{\infonce}^K(X;Y|f)\}, 
\eeq 
$\infonce$ implements the heuristic to discern {\it positive} samples from the {\it negative} samples, whereas the {\it positives} from the joint data distribution $p(x,y)$, and the {\it negatives} are randomly paired samples from respective marginal distributions $p(x)p(y)$. Here $f(x,y)>0$ is known as the {\it critic function} and we have used $p^K(x,y)$ to denote $K$ independent draws, which is the {\it sample size} or {\it (mini) batch-size}. Note we have used $\oplus$ and $\ominus$ to emphasize positive and negative samples. Mathematically, $\infonce$ constructs a formal lower bound to the mutual information, given by the following statement:
%Intuitively, $\infonce$ tries to discern the {\it positive} samples from the {\it negative} samples to maximize the knowledge of information. 
\begin{prop}
\label{thm:infonce}
\texttt{InfoNCE} is an asymptotically tight lower bound to the mutual information, {\it i.e.}, 
\beq
I(X;Y) \geq I_{\infonce}^K(X;Y|f), \quad \lim_{K\rightarrow\infty} I_{\infonce}^K(X;Y) \rightarrow I(X;Y). 
\eeq
\end{prop}
\vspace{-10pt}
A few technical remarks are useful for our subsequent developments: ($i$) the $K$-sample $\infonce$ estimator is upper bounded by $\log K$; ($ii$) in practice, $\infonce$ is implemented with the \texttt{CrossEntropy} loss for multi-class classification, where $f(x,y)$ is parameterized by its logit $g_{\theta}(x,y)=\log f(x,y)$; ($iii$) optimizing for $f(x,y)$ tightens the bound, and the bound is sharp if $f(x,y) = p(x|y) e^{c(x)}$, where $c(x)$ is an arbitrary function on $\CX$; and ($iv$) $\infonce$'s successes have been largely credited to the fact that its empirical estimator has much smaller variance relative to competing solutions. More details on these technical points are either expanded in later sections or deferred to the Appendix.

\vspace{-3pt}
\subsection{InfoNCE for contrastive representation learning}
\vspace{-3pt}

MI estimation is keenly connected to the literature on contrastive representation learning, with prominent examples such as \texttt{Word2vec} \citep{mikolov2013distributed}, \texttt{MoCo} \citep{he2020momentum}, \texttt{SimCLR} \citep{chen2020simple}, \texttt{SimCSE} \citep{gao2021simcse}, {\it etc}. In such works, one wants to build a robust feature extractor $\Enc(x): \CX \rightarrow \CZ$ in an unsupervised way, such that the encoded representation $z=\Enc(x)$ robustly encodes useful information in $x$ for various downstream applications.  A general framework adopted by such works is that one first specifies a set of valid data augmentation transformations $\CT\triangleq \{ t: \CX \rightarrow \CX \}$ ({\it e.g.}, for image data these transformations typically include crop, resize, flip, color distortion, cutout, noise corruption, blurring, filtering, {\it etc}.). Then, one tries to optimize for the heuristic goal that the encoded representation for the same data with different augmentations, sometimes referred to as different {\it views} \citep{tian2019contrastive}, should still be more {\it similar} compared to those encoded from different data points ({\it i.e.}, negative samples). Dot-product $\langle a, b\rangle = a^T b$ is often employed for efficient evaluation of similarity, and representations are typically normalized to unit spherical spaces ({\it i.e.}, $\| z \| = 1$) \citep{bojanowski2017unsupervised}. More specifically, let $t,t' \in \CT$ be two different transforms, these representation learning objectives optimize for different variants of the following cross-entropy-like loss %\zhe{be careful? eqn 1 has K samples in total; eqn 3 here has K+1 samples in total? not consistent.}\cytao{That's something I also want to discuss. Using $K+1$ makes the notation a bit clumsy.}
\beq
\ell(z, z_{\oplus}', \{z_{\ominus,k}'\}_{k=1}^K) = - \log \frac{\exp(\beta\langle z, z_{\oplus}'\rangle)}{\sum_{z'\in \{\oplus, \ominus\}} \exp(\beta\langle z, z'\rangle)},
\label{eq:nce}
\eeq
where $z = \Enc(t(x)), z_{\oplus}' = \Enc(t'(x)), z_{\ominus,k}' = \Enc(t'(x_k))$ for $x_k \neq x$, and $\beta>0$ is the inverse temperature parameter. %\cytao{As clarified in Sec 3.3, $\beta$ is not a hyper-parameter.}. 
Note (\ref{eq:nce}) essentially tries to predict the positive sample out of the negative samples. One can readily recognize that (\ref{eq:nce}) is mathematically equivalent to the \texttt{InfoNCE} target (\ref{eq:infonce}), up to a constant term $\log K$ and a much simplified bi-linear critic function $g(z, z') = \beta\langle z, z' \rangle$. This implies contrastive objectives are essentially optimizing for the mutual information $I(t(X);t'(X))$ between different views of the data. Let $z = \Enc(t(x))$ and $z'=\Enc(t'(x))$, by the data processing inequality, we have $I(t(X);t'(X)) \geq I(Z;Z') \geq I_{\infonce}(Z;Z')$. 
%\beq
%I(Z;Z')
%\eeq
A general observation made in the contrastive representation learning literature is that performance improves as $K$ gets larger, {\it e.g.}, in \texttt{SimCLR} the performance grows with $K$ and reaches its optimum at $K=4,096$ with a \texttt{ResNet50} architecture. These observations are consistent with our theoretical assertion that this MI bound gets tight as $K\rightarrow \infty$. However, this fact is challenging for budgeted applications and many investigators simply because training at such a scale is economically unaffordable. 

\vspace{-3pt}
\subsection{The failure of InfoNCE with small batch-sizes}
\vspace{-3pt}
\label{sec:nce_fail}

Despite $\infonce$'s sweeping successes, here we provide a careful analysis to expose its {\it Achilles heel}. Particularly, we reveal that as the $\infonce$ estimate approaches saturation ({\it i.e.}, $\hat{I}_{\infonce} \rightarrow \log K$, where $\hat{I}$ denotes an empirical sample estimate), its learning efficiency plunges due to limited numerical precision, hindering further improvements. This clarifies $\infonce$'s small-sample collapse and motivates our repairs in subsequent sections. 

%This not only clarifies why $\infonce$ works unsatisfactorily for small batch-sizes, but also motivates our development of repairs in subsequent sections to overcome this limitation. 

Recall in $\infonce$ that the loss is computed from the \texttt{CrossEntropy} loss. For most deep learning platforms, the internal implementations exploit the stable \texttt{logsumexp} trick 
\beq
\ell_{\text{CE}} = \texttt{logsumexp}(\{g_{ij}^{\ominus}\}) - g_{ii}^{\oplus} = \{ \log (\sum\nolimits_{j \in \{\oplus, \ominus\}} \exp(g_{ij}^{\ominus} - g_{\max}^{\ominus})) + g_{\max}^{\ominus} \} - g_{ii}^{\oplus} , 
\eeq
to avoid numerical overflow, where $g_{ij}\triangleq g_{\theta}(x_i, y_j)$. With a powerful learner for $g_{\theta}(x,y)$ and a small $K$ such that $I(X;Y)>\log K$, we can reasonably expect $\hat{I}_{\infonce} \approx \log K$ after a few training epochs. 
%Now let us assume we have a powerful learner for $g_{\theta}(x,y)$ but only limited computation budget ({\it i.e.}, $\log K< I(X;Y)$).  So after a number of training epochs, we can reasonably expect $\hat{I}_{\infonce} \approx \log K$. 
Since $g_{ii}^{\oplus}$ itself is also included in the negative samples, this implies $g_{ii}^{\oplus} \gg g_{ij}^{\ominus} ,\,\, \forall j\neq i$ almost always holds true, because $\hat{I}_{\infonce} = \log \frac{\exp(g_{ii}^{\oplus})}{\frac{1}{K}\sum_j \exp(g_{ij}^{\ominus})} \approx \log \frac{\exp(g_{ii}^{\oplus})}{\frac{1}{K}\exp(g_{ii}^{\oplus})} = \log K$. 
%\zhe{confusing.} \cytao{Since $g_{ii} \gg g_{ij}$, so $\exp(g_{ii})$ dominates the denominator.}
So the contrast now becomes 
\beq
\ell_{\text{CE}} = g_{ii}^{\oplus} + \log (\sum\nolimits_j \exp(g_{ij}^{\ominus} - g_{ii}^{\oplus})) - g_{ii}^{\oplus} = g_{ii}^{\oplus} + \log (1+o(1)) - g_{ii}^{\oplus} \approx 0 .
\eeq
This is where the algorithm becomes problematic: for low-precision floating-point arithmetic, {\it e.g.}, \texttt{float32} or \texttt{float16} as in standard deep learning applications, the relative error is large when two similar numbers are  subtracted from one another. The contrastive terms $g_{ij}^{\ominus} - g_{ii}^{\oplus}$, which are actually contributing the learning signals, will be engulfed by the dominating $g_{ii}^{\oplus}$ and accumulate rounding errors. For an easy fix, one can explicitly modify the computation graph to cancel out $g_{ii}^{\oplus}$, as the \texttt{CrossEntropy} shipped with platforms such as \texttt{Tensorflow} and \texttt{PyTorch} does not do so.

%While in theory, this can be avoided by rearranging the computation graph, so that the identical items are explicitly canceled before interacting with other contributing terms. Unfortunately, that is a not the default behavior of the off-the-shelf \texttt{auto-grad} routines employed by deep learning platforms such as \texttt{Tensorflow} or \texttt{PyTorch}. And consequently, the numerical errors accumulate during loss, and in term, gradient computation, as the \texttt{auto-grad} proceeds in-order to collect and aggregate loss and gradients. Now we see what happens when $\infonce$ has been trained to max-out its representation capacity, thereby impeding further performance improvements. 

%$g_{ii}^{\oplus} > g_{ij}^{\ominus} ,\,\, \forall j\neq i$. This means $g_{\max}^{\ominus} = g_{ii}^{\oplus}$ holds true with large probability. If the estimate is close to the maximal capacity of $K$-samples ({\it i.e.}, $\log K$), then we must have $g_{ii}^{\oplus} \gg g_{ij}^{\ominus} ,\,\, \forall j\neq i$. 
%\beq
%\ell_{\text{CE}} = \texttt{logsumexp}(\{g_{ij}^{\ominus}\}) - g_{ii}^{\oplus} = g_{\max}^{\ominus} - g_{ii}^{\oplus} + \log (\sum_j \exp(g_{ij}^{\ominus} - g_{\max}^{\ominus})) 
%\eeq
%where $g_{\max}^{\ominus} \triangleq \max_j\{ \{g_{ij}^{\ominus}\} \}$. As we can see, this objective explicitly contrasts the positive affinity score with the ``hardest'' negative affinity score $g_{\max}^{\ominus}$. Loss of learning efficiency occurs if $g_{ii}^{\oplus} \approx g_{\max}^{\ominus}$

\vspace{-8pt}
\section{FlatNCE And Generalized Contrastive Representation Learning}
\vspace{-6pt}

%riveting
%by and large

In this section, we deconstruct the building principles of contrastive learning to make fixes to the $\infonce$. Our new proposal, named $\FLAT$, addresses the limitations of the na\"ive $\infonce$ and comes with strong theoretical grounding. Detailed derivations are relegated to the Appendix.

%We start from very simple heuristics and intuitively derive a surrogate objective we call the $\FLAT$, that overcomes the limitations of the na\"ive $\infonce$ objective. Then we show $\FLAT$ actually enjoys strong theoretical groundings, and can be generalized to a novel flexible contrastive learning framework. To save space, details derivations are relegated to the Appendix.

\begin{wrapfigure}[11]{R}{0.53\textwidth}
\scalebox{0.9}{
\begin{minipage}{0.53\textwidth}
\vspace{-4em}
%\begin{figure}[t]
\begin{algorithm}[H]
%   \caption{Fenchel-InfoNCE}
\caption{$\FLAT$}
   \label{alg:flat}
\begin{algorithmic}
%\small
\STATE Empirical data distribution $\hat{p}_d = \{ (x_i, y_i) \}_{i=1}^n$ \\
[3pt]
%\STATE Initialize parameters $\theta, b$ \\
%\#\#\#\#\# Training \#\#\#\#\#
%\STATE $\BZ = \Enc$
\FOR{$t=1,2,\cdots$}
%\STATE Sample $(x_{i_t}, y_{i_t}) \sim \hat{p}_{d}(x,y)$, $y_{i_t'} \sim \hat{p}_d(y)$\\
\STATE Sample $i,i_k' \sim [1, \cdots, n], k'\in[1, \cdots, K]$ \\
[1pt]
%\STATE $\gv_{\oplus} = g_{\theta}(x_i, y_i) \in \BR^{m\times 1}$ \\
%\STATE $\gv_{\ominus} = g_{\theta}(x_i, y_{i_k'}) \in \BR^{m \times K}$\\
\STATE $\gv_{\oplus} = g_{\theta}(x_i, y_i), \gv_{\ominus} = g_{\theta}(x_i, y_{i_k'})$\\
[1pt]
\STATE \# $\texttt{logits} = [\gv_{\oplus}, \gv_{\ominus} ], \texttt{labels} = \bs{0}$ \\
%\STATE 
\STATE \# $\ell_{\infonce} = \texttt{CrossEntropy}( \texttt{logits}, \texttt{labels})$ \\
[1pt]
%\STATE \# $\ell_{\infonce} = \texttt{CrossEntropy}( [\gv_{\oplus}, \gv_{\ominus} ], \bs{0})$ \\
%\STATE $\texttt{clogits} = \texttt{logsumexp}(\gv_{\ominus} - \gv_{\oplus})$
\STATE $\texttt{clogits} = \texttt{logsumexp}(\gv_{\ominus} - \gv_{\oplus})$\\
[1pt]
%\STATE $\ell_{\FLAT} = \texttt{c}/\texttt{detach}[\texttt{c}], \texttt{c} = \exp(\texttt{clogits})$
\STATE $\ell_{\FLAT} = \exp(\texttt{clogits}-\texttt{detach}[\texttt{clogits}])$ \\
[1pt]
\STATE \# Use your favorite optimizer
%[1pt]
%\STATE $\CF = u + \exp(-u+g_{\ominus}-g_{\oplus})$\\
%[1pt]
%$\Psi_{t} = \Psi_{t} - \eta_t \nabla_{\Psi} \CF$
\ENDFOR
%\STATE {\bf function} $\infonce(\gv_{\oplus}, \gv_{\ominus})$
%\STATE  \hspace{1em} $\texttt{logits} = [\gv_{\oplus}, \gv_{\ominus} ], \texttt{labels} = \bs{0}$
%\STATE \hspace{1em} $\ell_{\infonce} = \texttt{CrossEntropy}( \texttt{logits}, \texttt{labels})$ \\
%\STATE {\bf end function}
%\STATE {\bf function} $\FLAT(\gv_{\oplus}, \gv_{\ominus})$
%\STATE  \hspace{1em} $\texttt{clogits} = \texttt{logsumexp}(\gv_{\ominus} - \gv_{\oplus})$
%\STATE \hspace{1em} $\ell_{\FLAT} = \exp(\texttt{clogits}-\texttt{detach}[\texttt{clogits}])$
%\STATE {\bf end function}
\end{algorithmic}
\end{algorithm}
\vskip -.05in
\vskip -.2in
\end{minipage}
}
\end{wrapfigure}

\vspace{-3pt}
\subsection{Making it flat: InfoNCE recasted}
\vspace{-3pt}

%Our analyses in Section \ref{sec:nce_fail} show why $\infonce$ fails in the small batch-size regime when sticking with the cross-entropy loss, so we walk back and see if there is any alternative heuristics we can leverage. 

Recall in the $\infonce$ objective the critic function $g_{\theta}(x,y)$ computes an affinity score for a pair of data $(x,y)$ and optimizes for the following intuition: paired $(x_i, y_i)$, known as the {\it positives}, should have larger affinity scores relative to the negative samples $\{ (x_i, y_j) \}$, where $y_j$ is randomly drawn from the marginal distribution of $y$. 
%In other words, $g_{\theta}(x_i, y_j) - g_{\theta}(x_i, y_i)<0$ is expected. When $\Delta_{ij} = -\infty \,\, \text{ for } i \neq j$, where $\Delta_{ij} \triangleq g_{\theta}(x_i, y_j) - g_{\theta}(x_i, y_i)$, 
%the $\infonce$ estimator attains its maximal value $\log K$. And this reveals a very useful heuristic: 
For a good representation, where the $\infonce$ estimation is maximized, we would like to make the affinity score contrasts $\Delta_{ij} \triangleq g_{\theta}(x_i, y_j) - g_{\theta}(x_i, y_i)$ as negative as possible. The challenge is to make negative contrasts continually improve after $I_\infonce$ has reached $\log K$.
%, then we break the spell casted on $\infonce$. 
%Instead of optimizing for the 

%Instead of , we directly model for their. 
%rather than targeting the positive-sample prediction accuracy, 
So motivated, we drop the positive-sample prediction accuracy, and instead seek to directly optimize the affinity score contrasts to make them dip further after $I_{\infonce}$ has saturated. We begin with a discussion of the desirable properties for such an objective: ($i$) {\it differential penalty} and ($ii$) {\it instance normalization}. By differential penalty, we want the objective to nonlinearly regularize the affinity contrast, {\it i.e.}, through the use of a {\it link function} $h(\Delta_{ij})$, such that it penalizes more heavily for smaller affinity differences, and has diminishing effect on affinity differences that are already sufficiently negative. Secondly, instance normalization attends to the fact that different $x$ may have different baselines for the affinity contrasts, which should be properly equalized. To see this, we recall in theory the optimal critic for $\infonce$ is $g^*(x,y) = \log p(x|y) + c(x)$, so the expected value of $\EE_{p(y')}[g(x,y')]$ varies with $x$, and it needs to be offset differently. 

With these design principles in mind, now we start to build such a surrogate objective. Naturally, the tempered exponential transform $\exp(\beta t)$ meets our expectation for the link function, where $\beta>0$ is a tuning parameter known as the {\it inverse temperature}. To simplify our discussion, we always treat $\beta=1$ unless otherwise noted. On the other hand, incentivized by $\infonce$'s successes due to small estimating variance, we want the resulting objective to enjoy a low-variance profile. Taking this to the extreme, we propose {\bf FlatNCE}, a {\it zero-variance} mini-batch MI optimizer, defined as
%\beq
%I_{\FLAT} = \texttt{detach}[I_{\infonce}] + \frac{\sum_{j} \exp(g_{\theta}(x_i, y_j) - g_{\theta}(x_i, y_i))}{\texttt{detach}[ \sum_{j} \exp(g_{\theta}(x_i, y_j) - g_{\theta}(x_i, y_i)) ]} - 1
%\eeq
\beq
I_{\FLAT} = \frac{\sum_{j} \exp(g_{\theta}(x_i, y_j') - g_{\theta}(x_i, y_i))}{\texttt{detach}[ \sum_{j'} \exp(g_{\theta}(x_i, y_{j'}') - g_{\theta}(x_i, y_i)) ]} , 
\label{eq:flat}
\eeq
where \texttt{detach[$f_{\theta}(x)$]} is an operation that bars gradient back-propagation. 
In accordance with the notation employed by $\infonce$, $j\in \{ 1, \cdots, K-1\}$ indexes the $K-1$ negative samples drawn from $p_d(y)$, so together with the positive sample (\ref{eq:flat}) gives a $K$-sample estimator. 
One may readily notice that this is a flat function, as $I_{\FLAT} \equiv 1$ for arbitrary inputs\footnote{Note that the gradient of $\FLAT$ is not flat, that is why we can still optimize the representation.}, which fulfills the zero-variance property. In this regard, we consider $\FLAT$ as a {\it self-normalized} contrastive objective. In the next section, we rigorously prove how $\FLAT$ is formally connected to $\infonce$, and why it is more preferable as confirmed by our experiments in Section \ref{sec:exp}.

%we want the objective to deform 

%design principles 
%
%Note that the individual affinity score needs to be properly aggregated, through an transform $h(t)$, otherwise one may encounter the degenerate solution, where some pairs of affinity score go to $-\infty$, while the others may head the other way, while their sum keeps decreasing. This transform should be a monotonic function, and it should penalize more heavily for non-negative affinity differences, and has diminishing effect on affinity difference that is already sufficiently negative. One proper aggregation function is the exponential transform $\exp(t/\tau)$, where $\tau>0$ is a tuning parameter. Similarly, different $x$ may have different baseline affinities.  in theory $g^*(x,y) = \log p(x|y) + c(x)$, so the expected value of $\EE_{p(y')}[g(x,y')]$ varies with $x$ and needs to be offset properly. 
%Also, we need to take a 

%\beq
%\sum_{i} \left\{ \frac{\sum_{j}  h(\Delta_{ij})}{s(x_i)} \right\}, \,\, \Delta_{ij} = g_{\theta}(x_i, y_j) - g_{\theta}(x_i, y_i) ,
%\eeq
%where $s(x)$ is the scaling function and $h(u)$ is the aggregation function. Without loss of generality, we assume $h(u)>0$ and it is monotonically increasing. 

\vspace{-3pt}
\subsection{Understanding FlatNCE}
\vspace{-3pt}

We first define the following variant of the $\FLAT$
\beq
\label{eq:plus}
I_{\FLAT}^{\oplus}(g_{\theta}) = \frac{1+\sum_{j} \exp(g_{\theta}(x_i, y_j') - g_{\theta}(x_i, y_i))}{1+\texttt{detach}[ \sum_{j'} \exp(g_{\theta}(x_i, y_{j'}') - g_{\theta}(x_i, y_i)) ]} .
\eeq
Note that $I_{\FLAT}^{\oplus}(g_{\theta})$ corresponds to adding the positive sample $y_i$ to the set of negative samples, because the zero contrast of the positive sample always gives the constant one. The following statement verifies $I_{\FLAT}^{\oplus}(g_{\theta})$ is equivalent to $\infonce$ in terms of differentiable optimization. 
%Readers may have readily noticed that taking the gradient wrt the model parameters for $I_{\FLAT}^{\oplus}(g_{\theta})$ recovers the gradient of $\infonce$:
%Readers may have readily noticed that $I_{\FLAT}^{\oplus}(g_{\theta})$ is mathematically equivalent to $\infonce$ in terms of differentiable optimization, summarized by the following assertion.
\begin{prop}
\label{thm:equiv} $\nabla_{\theta} I_{\FLAT}^{\oplus}(g_{\theta}) = \nabla_{\theta} I_{\infonce}(g_{\theta})$. 
%\beq
%\nabla_{\theta} I_{\FLAT}^{\oplus}(g_{\theta}) = \nabla_{\theta} I_{\infonce}(g_{\theta})
%\eeq
\end{prop}
% Equation (\ref{eq:plus}) and Proposition \ref{thm:equiv} remind us the discussion we had in Section \ref{sec:nce_fail} on the intuitions why $\infonce$ is failing for small batch-sizes. Now we extend that discussion by showing the gradient of $\FLAT$ and its variant (so equivalently, $\infonce$) is given by a self-normalized importance-weighted gradient estimator, formalized in the Proposition below.
Equation (\ref{eq:plus}) and Proposition \ref{thm:equiv} indicate how $\infonce$ and $\FLAT$ are connected. Based on this, we can continue our discussion in Section \ref{sec:nce_fail} on why $\infonce$ fails for small batch sizes. We start by showing the gradient of $\FLAT$ and its variant $I_{\FLAT}^{\oplus}$ (so equivalently, $\infonce$) is given by a self-normalized importance-weighted gradient estimator, as formalized below.
\begin{prop}
\label{thm:iw}
The gradient of $\FLAT$ is an importance-weighted estimator of the form
\beq
\nabla I_{\FLAT} = \sum\nolimits_j w_j \nabla g_{\theta}(x_i, y_j) - \nabla g_{\theta}(x_i, y_i), \quad \text{ where } w_j = \frac{\exp(g_{\theta}(x_i, y_j'))}{ \sum_{j'}  \exp(g_{\theta}(x_i, y_{j'}')))}. 
%\vspace{-5pt}
\eeq
\end{prop}
\vspace{-8pt}
% Note the above result applies both $I_{\FLAT}$ and $I_{\FLAT}^{\oplus}$.
Without loss of generality, let us denote $y_0' = y_i$, so when $\hat{I}_{\infonce}$ approaches $\log K$, we know $w_0\approx 1, w_{j>0} \approx 0$, and consequently $\nabla I_{\infonce} \approx \nabla g_{\theta}(x_i, y_i) - \nabla g_{\theta}(x_i, y_i) = 0$. 
%Without loss of generality let us denote $y_0' = y_i$. So when $\hat{I}_{\infonce}$ approaches $\log K$, we will have $w_0\approx 1, w_{j>0} \approx 0$, and thus $\nabla I_{\infonce} \approx \nabla g_{\theta}(x_i, y_i) - \nabla g_{\theta}(x_i, y_i) = 0$. 
Consequently, as long as the positive sample is in the denominator the learning signal vanishes. What makes matters worse, the low-precision computations employed to speed up training introduce rounding errors, further corrupting the already weak gradient. On the other hand, in $\FLAT$ larger weights will be assigned to the more challenging negative samples in the batch, thus prioritizing hard negatives. 
%{\bf \textcolor{blue}{ (TAO: Equilibrium maybe established, to prove.) } }

Proposition \ref{thm:iw} also sheds insights on temperature annealing. Setting $\beta \neq 1$ re-normalizes the weights by exponential scaling ({\it i.e.}, $w_j(\beta) = w_j^{\beta}/\sum_{j'} w_{j'}^{\beta}$). So the optimizer will focus more on the hard negative samples at a lower temperature ({\it i.e.}, larger $\beta$), while for a higher temperature it treats all negative samples more equally. This new gradient interpretation reveals that $\beta$ affects the learning dynamics in addition to the well-known fact that $\beta$ modulates MI bound tightness.

Lastly, to fill in an important missing piece, we prove that $\FLAT$ is a formal MI lower bound. 

\begin{lem}
For $\{ (x_j, y_j) \}_{j=1}^K$, let $I_{\infonce}^K(g_{\theta}) \triangleq - \log \frac{1}{K}\sum_{j} \exp(g_{\theta}(x_{1}, y_{j}^{\ominus}) - g_{\theta}(x_{1}, y_{1}^{\oplus}))$. Then for arbitrary $u\in \BR$, we have inequality
\beq
I_{\infonce}^K(g_{\theta}) \geq 1 -  u - \frac{1}{K}\sum\nolimits_{j} \exp(-u + g_{\theta}(x_1, y_{j})-g_{\theta}(x_1, y_1)),
\label{eq:dual}
\eeq
and the equality holds when $u = \frac{1}{K}\sum_j \exp(g_{\theta}(x_1, y_{j})-g_{\theta}(x_1, y_1))$. 
\end{lem}

What makes this particularly interesting is that $\FLAT$ can be considered the conjugate dual of $\infonce$. In convex analysis, $u$ and $g$ in (\ref{eq:dual}) are known as the {\it Fenchel conjugate pair} \citep{fenchel1949conjugate,tao2019fenchel, guo2021tight}. By taking the expectation wrt $p^K(x,y)$ and setting $u(\{ (x_j, y_j) \})$ to its optimal value, we essentially recover $I_{\FLAT}^{\oplus}(g_{\theta})$: the only difference to the conjugate of $\infonce$ is the term $(1-u)$ which is considered fixed and does not participate in optimization. As such, the following Corollary is immediate\footnote{Using a similar technique, we can also show (\ref{eq:flat}) lower bounds mutual information. Details in Appendix.}.
%we prove $\FLAT$ is a lower bound to mutual information. 

\begin{col}
$I_{\FLAT}^{\oplus,K}(g_{\theta}) = I_{\infonce}^K(g_{\theta}) \leq I(X;Y), \,\,\, I_{\FLAT}^{K}(g_{\theta})  \leq I(X;Y)$. 
\end{col}

% {\it Remark.} Using a similar technique, we can also show (\ref{eq:flat}) lower bounds mutual information. The proof is a little bit more involved, but the basic idea is to simply take a $K'>K$, rearrange the terms and then show $I_{\FLAT}^K(g_{\theta}) \leq I_{\infonce}^{K'}(g_{\theta}) \leq I(X;Y)$. See Appendix for details. 

\vspace{-5pt}
\subsection{Generalizing FlatNCE}
\vspace{-3pt}

The formulation of $\FLAT$ enables new possibilities for extending contrastive representation learning beyond its original form. In this section, we discuss some generalizations that make contrastive learning more flexible, including new tools for training diagnosis and tuning.  

{\bf Effective sample-size (ESS) scheduling.} While existing contrastive training schemes view temperature parameter $\beta$ as a static hyper-parameter that tunes model performance, our Proposition \ref{thm:iw} shows that it also plays a dynamic role in updating the critic $g_{\theta}(x, y)$. This motivates us to anneal the contrastive learning by scheduling the temperature parameter. To exact better control over the training process through $\beta$, we appeal to the notation of normalized {\it effective sample-size} (ESS), defined by
\beq
\ESS \triangleq 1/\mathlarger\{K\sum\nolimits_j w_j^2\mathlarger\} \in [1/K, 1].
\eeq
$\ESS$ provides richer information about the training than the estimated MI. A value close to $\ESS\approx 1$ implies gradient diversity, as all samples are contributing to the gradient equally; $\ESS\approx 0$ raises concern, as a small fraction of samples dominate the gradient, thus leading to a higher variance and consequently unstable training, as in the case of $\infonce$. 
So rather than directly employing temperature scheduling, we can instead instruct the model to train with the targeted level of $\ESS$ at each stage of training, namely $\ESS$ scheduling, which adaptively adjusts the temperature for the current model (see Algorithm S1 in Appendix). In early training it is more beneficial to aim at a higher $\ESS$, such that model can more efficiently assimilate knowledge from a larger sample pool. As training progresses we gradually relax $\ESS$ constraints to allow the model to reach tighter $\MI$ bounds, while making sure gradient variance are kept under control.

{\bf H\"older FlatNCE.} To further generalize contrastive learning, we re-examine the objective of $\FLAT$. A key observation is that the numerator aggregates individual {\it evidence} of MI from the negative samples $(x,y')\sim p(x)p(y')$ through the critic function $g_{\theta}(x,y)$, with arithmetic mean. Possibilities are that if we change the aggregation step, we also change how it learns MI in a way similar to the importance weighting perspective discussed above. This inspires us to consider the more general aggregation procedures, such as the H\"older mean defined below.
%In order to generalize contrastive learning, we consider the more general H\"older averaging operation defined below. 
\begin{defn}[H\"older mean] 
For $\{a_i \in \BR_+\}_{i=1^n}$ and $\gamma \in \BR$, the H\"older mean is defined as 
$
m_\gamma(\{ a_i \}_{i=1}^n) = \left(\frac{1}{n}\sum_i a_i^\gamma \right)^{\frac{1}{\gamma}}
$.
%$ \CM_{\alpha}(a,b;\beta) = \left[\beta a^\alpha+(1-\beta)b^{\alpha}\right]^{\frac{1}{\alpha}}$
%, and for $\alpha=0$ we use $\CM_{0} = \lim_{\alpha \rightarrow 0} \CM_{\alpha}$.
\end{defn}
Note H\"older mean recovers many common information pooling operations, such as $\min$ ($\gamma=-\infty$), $\max$ ($\gamma=\infty$),  geometric mean ($\gamma\rightarrow 0$), root mean square ($\gamma=2$), and arithmetic mean ($\gamma=1$) as employed in our $\FLAT$. This allows us to define a new family of contrastive learning objectives.
%$\gamma = -\infty \Rightarrow \min$, $\gamma = \infty \max$, $\gamma = 0 \Rightarrow$ geometric mean, $\gamma = 1 \Rightarrow $ arithmetic mean, $\gamma=2 \Rightarrow $ root mean square.   
%\hl{In Figure S1 in the SM, we visualize such H\"older paths, with special examples at $\alpha=0, 1$ respectively recovering the geometric and arithmetic means.} 
%{See Figure S1 in the SM for examples of H\"older paths.}
%We can analogously define the thermodynamic curves wrt H\"older paths, with the following statement directly generalizing the monotonicity result.
\begin{defn}[H\"older-$\FLAT$]
$
I_{\gamma} \triangleq \sum_i \frac{m_{\gamma}(\{ \exp(g_{ij}^{\ominus} - g_{ii}^{\oplus}) \}_j)}{\texttt{detach}[m_{\gamma}(\{ \exp(g_{ij}^{\ominus} - g_{ii}^{\oplus}) \}_j)]}
$.
\end{defn}
The following Proposition shows that H\"older-$\FLAT$ is equivalent to annealed $\FLAT$. 
\begin{prop}
$\nabla I_{\gamma}(g_{\theta}) = \nabla I_{\FLAT}(\gamma \cdot g_{\theta})$.
\end{prop}
As an important remark, we note the sample gradient of $\FLAT$ is a (randomly) re-scaled copy of the true gradient (normalized by $Z_{\theta}$ instead of $\hat{Z}_{\theta}$), so we are still optimizing the model in the right direction using stochastic gradient descent (SGD) \citep{robbins1951stochastic}. This property can be used to ascertain the algorithmic convergence of $\FLAT$, formalized in the Proposition below. Details in Appendix.
% Note that the sample gradient is a randomly rescaled copy of the true gradient (normalized by $Z_{\theta}$ instead of $\hat{Z}_{\theta}$), so we are still optimizing the model in the right direction using SGD \citep{robbins1951stochastic}. This property can be used to ascertain the algorithmic convergence of $\FLAT$, formalized in the Proposition below. See our Appendix for detailed proof and technical conditions needed.
\begin{prop}[Convergence of $\FLAT$, simple version]
\label{thm:conv}
Under the technical conditions in Assumption A1, with Algorithm \ref{alg:flat} 
$\theta_t$ converges in probability to a stationary point of the unnormalized mutual information estimator $I(\theta) \triangleq \EE_{p(x,y)}[g_{\theta}(x,y)] - \EE_{p(x)}[\log Z_{\theta}(x)]$ ({\it i.e.}, $\lim_{t\rightarrow\infty}\| \nabla I(g_{\theta_t}) \|=0$), where $ Z_{\theta}(x) \triangleq \EE_{p(y)}[e^{g_{\theta}(x,y)}]$. Further assume $I(\theta)$ is convex with respect to $\theta$, then $\theta_t$ converges  in probability to the global optimum $\theta^*$ of $I(\theta)$.
\end{prop}

\vspace{-8pt}
\section{Rethinking Contrastive Learning: Experimental Evidence \& Discussions}
\label{sec:discuss}
\vspace{-5pt}

% \begin{figure}[t!]
% \centering
% \begin{minipage}{.48\textwidth}
%   \centering
%   \includegraphics[width=1.\textwidth]{figures/results/batchsize-cifar.pdf}
%   \vspace{-1.5em}
%   \captionof{figure}{Comparison of sample efficiency between $\infonce$ and $\FLAT$ on \texttt{Cifar}. \label{fig:batchsize}}
% %   Solid lines are $\infonce$ and $+$-noted lines are $\FLAT$. $\FLAT$ reached better performance using only a fraction of the mini-batch size compare to $\infonce$. 
% %   \label{fig:test1}
% \end{minipage}%
% \hspace{5pt}
% \begin{minipage}{.48\textwidth}
%   \centering
%   \includegraphics[width=1.\textwidth]{figures/results/speedup.pdf}
%   \vspace{-1.5em}
%   \captionof{figure}{Speed up of large batch training. \label{fig:speedup} More results see Appendix.}
% %   \label{fig:test2}
% \end{minipage}
% \vspace{-1.em}
% \end{figure}

\begin{figure}[t!]
\centering
\begin{minipage}{.355\textwidth}
  \centering
  \includegraphics[width=1.\textwidth]{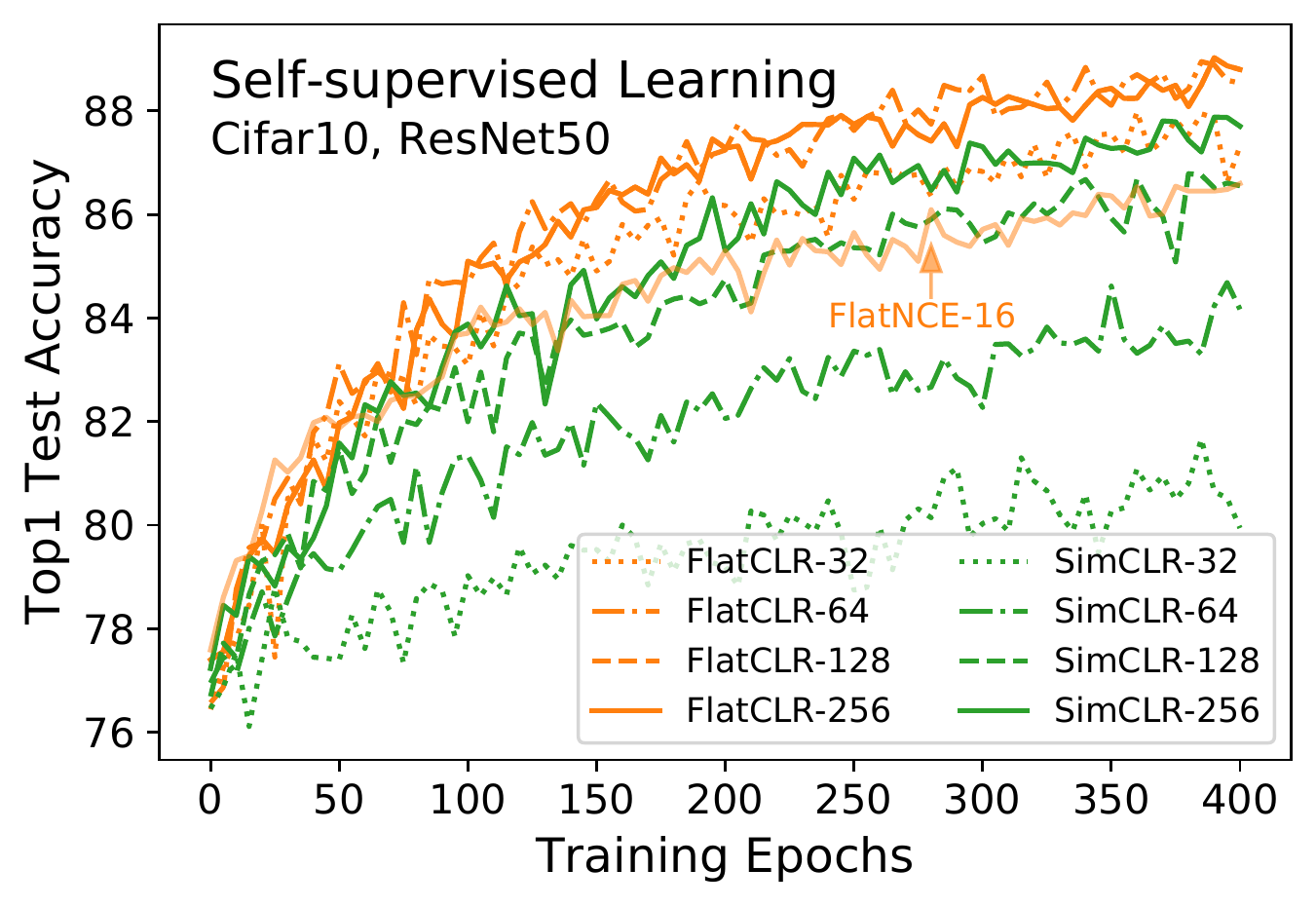}
  \vspace{-1.5em}
  \captionof{figure}{Sample efficiency comparison for $\SimCLR$ and $\FCLR$ on \texttt{Cifar10}. \label{fig:batchsize}}
%   Solid lines are $\infonce$ and $+$-noted lines are $\FLAT$. $\FLAT$ reached better performance using only a fraction of the mini-batch size compare to $\infonce$. 
%   \label{fig:test1}
\end{minipage}%
\hspace{2pt}
\begin{minipage}{.355\textwidth}
%\vspace{1em}
  \centering
  \includegraphics[width=1.\textwidth]{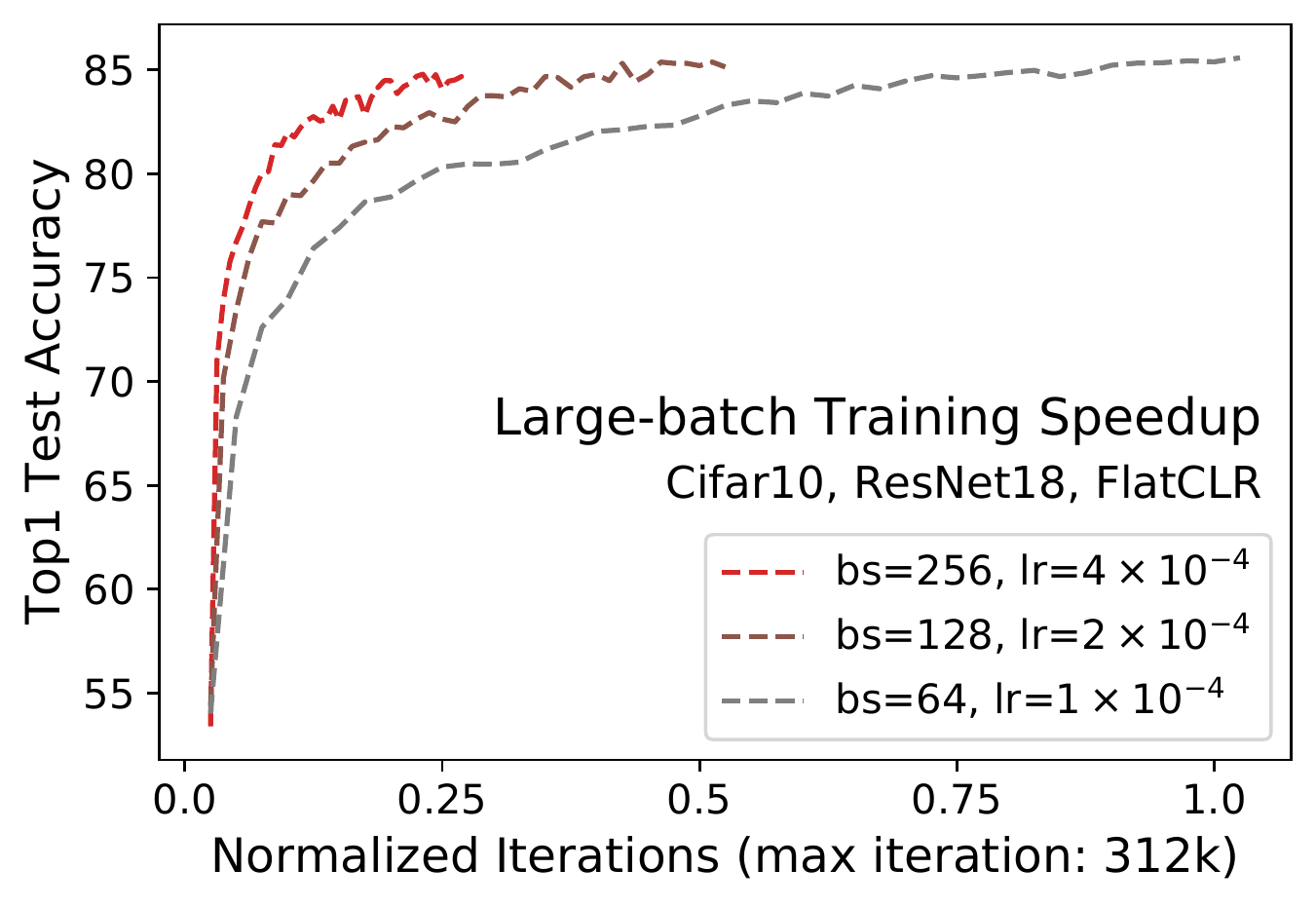}
  \vspace{-1.5em}
  \captionof{figure}{Speed up of large-batch training.  Larger batch leads to faster convergence. \label{fig:speedup} }
%   \label{fig:test2}
\end{minipage}
\hspace{2pt}
\begin{minipage}{.25\textwidth}
% \vspace{1.7em}
  \centering
  \includegraphics[width=1.\textwidth]{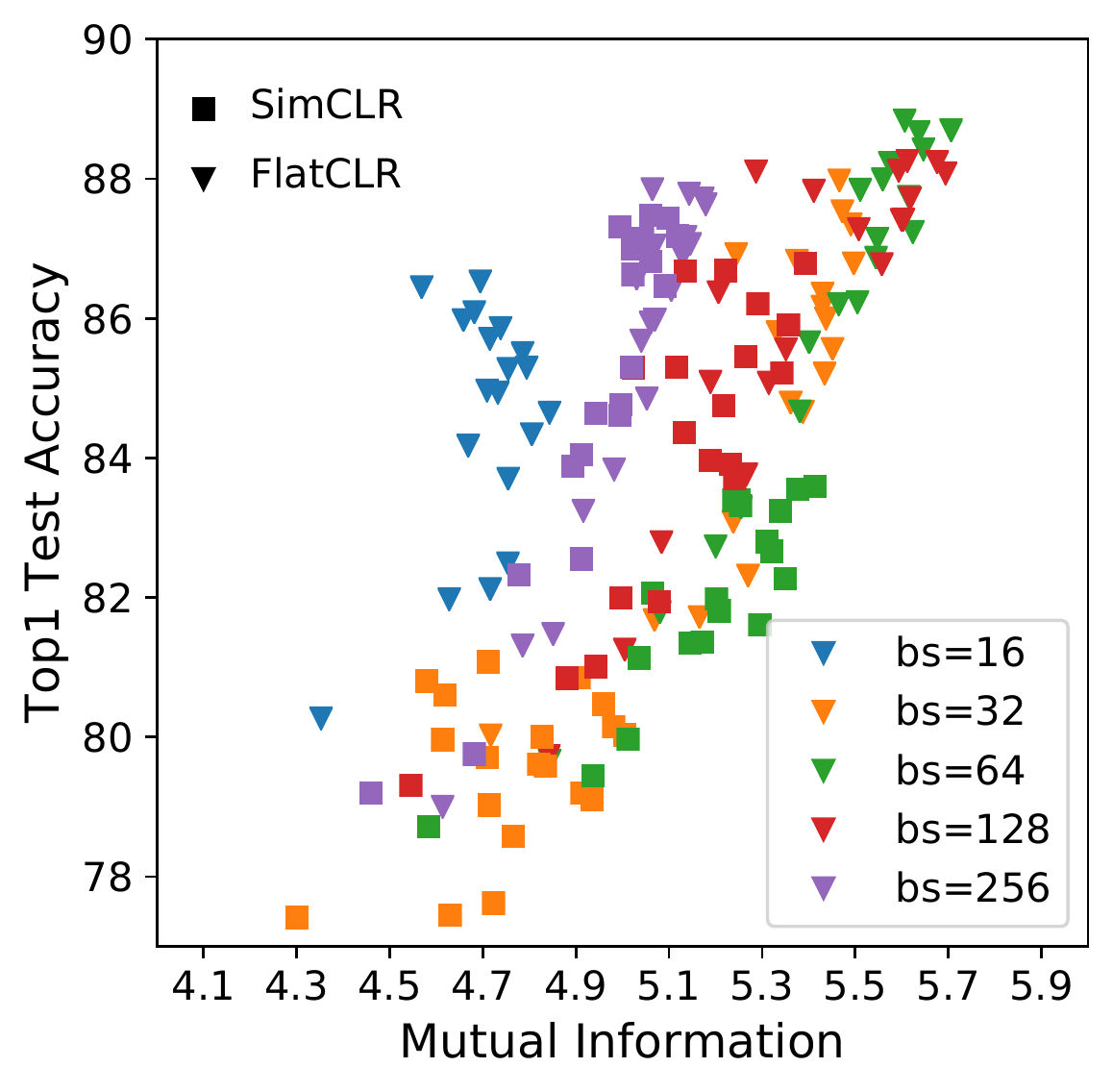}
  \vspace{-1.5em}
\captionof{figure}{Representation MI strongly correlates with performance. \label{fig:mi_acc}}
%   \label{fig:test1}
\end{minipage}%
\vspace{-1.em}
\end{figure}

% \begin{figure}[t!]
% \begin{center}
% %\includegraphics[width=.7\textwidth]{figures/results/cifar-flo-sim}
% \includegraphics[width=.48\textwidth]{figures/results/batchsize-cifar}
% \includegraphics[width=.48\textwidth]{figures/results/speedup}
% \end{center}
% \vspace{-1em}
% \caption{(a) Comparison of sample efficiency between $\infonce$ and $\FLAT$ on \texttt{Cifar}. Solid lines are $\infonce$ and $+$-noted lines are $\FLAT$. $\FLAT$ reached better performance using only a fraction of the mini-batch size compare to $\infonce$. \label{fig:batchsize} (b) Speed up of large batch training. \label{fig:speedup} More results see Appendix. }
% \vspace{-1.em}
% \end{figure}

% We devote this section to the activate discussion
We contribute this section to the active discussions on some of the most important topics in contrastive learning. 
%With our theory and new experimental evidence, we debunk misbeliefs and clarify confusions, and importantly, provide practical guidelines and raise new questions for future investigations. 
Our discussions will be grounded on the new experimental results from \texttt{Cifar10} with a \texttt{ResNet} backbone, with a \texttt{PyTorch} codebase of the $\infonce$-backed \texttt{SimCLR} and its $\FLAT$ counterpart $\FCLR$. Note instead of trying to set new performance records (because of limited computational resources in our university setting), experiments in this section are designed to reveal important aspects of contrastive learning, and to ensure our results can be easily reproduced with reasonable computation resources. Details of our setups are elaborated on in Section \ref{sec:exp}.

% \subsection{Self-normalized contrastive learning as constrained optimization}

% \begin{figure}[t!]
% \begin{center}
% %\includegraphics[width=.7\textwidth]{figures/results/cifar-flo-sim}
% \includegraphics[width=.7\textwidth]{figures/results/speedup}
% \end{center}
% \caption{Speed up of large batch training. \label{fig:speedup}}
% \end{figure}

% \begin{wrapfigure}[12]{L}{0.4\textwidth}
% % \vspace{-2.8em}
% % \scalebox{1.}{
% %\hspace{-.5em}
% \begin{minipage}{1.\textwidth}
% \begin{figure}[H]
% \begin{center}{
% \includegraphics[width=.4\textwidth]{figures/results/mi_acc}
% }
% % \vspace{-2em}
% \caption{\small Self-MI of the learned representation strongly correlates with downstream performance. \label{fig:mi_acc}}
% \end{center}
% % updated policy sampler
% \vspace{-1.5em}
% \end{figure}
% \end{minipage}
% % }
% \end{wrapfigure}

{\bf Breaking the curse, small-batch contrastive learning revived.} We show that with our novel $\FLAT$ objective, successful contrastive learning applications are no longer exclusive to the costly large-batch training. In Figure \ref{fig:batchsize} we see pronounced small-sample performance degradation for $\SimCLR$, while the $\FCLR$ is far less sensitive to the choice of batch size. In fact, we see $\FCLR$-$16$ matches performance of its $\SimCLR$-$128$ counterpart, corresponding to an $8\times$ boost in efficiency. And in all cases $\FCLR$ consistently works better compared to the same-batch-size $\SimCLR$. Despite the encouraging improvements in the small-batch regime, large-batch training does provide better results for both $\SimCLR$ and $\FCLR$. Additionally, leveling up parallelism greatly reduces the overall training time (Figure \ref{fig:speedup}), as a larger batch-size enables stable training with a larger learning rate \citep{bottou2010large, keskar2016large, goyal2017accurate} \footnote{While learning rate scheduling does affect performance, it is beyond the scope of our current investigation.}. The main merits of our result are: ($i$) the enabling of contrastive learning for very budgeted applications, where large-batch learning is prohibitive; and ($ii$) consistent improvement over $\infonce$, especially wrt the cost-performance trade-off.

% We see that the $\infonce$-based $\SimCLR$ greatly suffered from the small batch-size, as the performance dropped $7.5\%$ when the batch size is reduced from $128$ to $16$. In comparison, $\FLAT$ dropped $5\%$ for bs=$16$, and only $2\%$ for bs=$32$. Perhaps surprisingly, $\FLAT$ with bs=$32$ $\infonce$ with bs=$128$ by $2\%$. 

\begin{figure}[t!]
\centering
% \begin{minipage}{.48\textwidth}
%   \centering
%   \includegraphics[width=1.\textwidth]{figures/results/mi_acc.pdf}
%   \vspace{-1.5em}
% \captionof{figure}{Self-MI of the learned representation strongly correlates with downstream performance. \label{fig:mi_acc}}
% %   \label{fig:test1}
% \end{minipage}%
% \hspace{5pt}
\begin{minipage}{.42\textwidth}
\vspace{.2em}
  \centering
  \includegraphics[width=1.\textwidth]{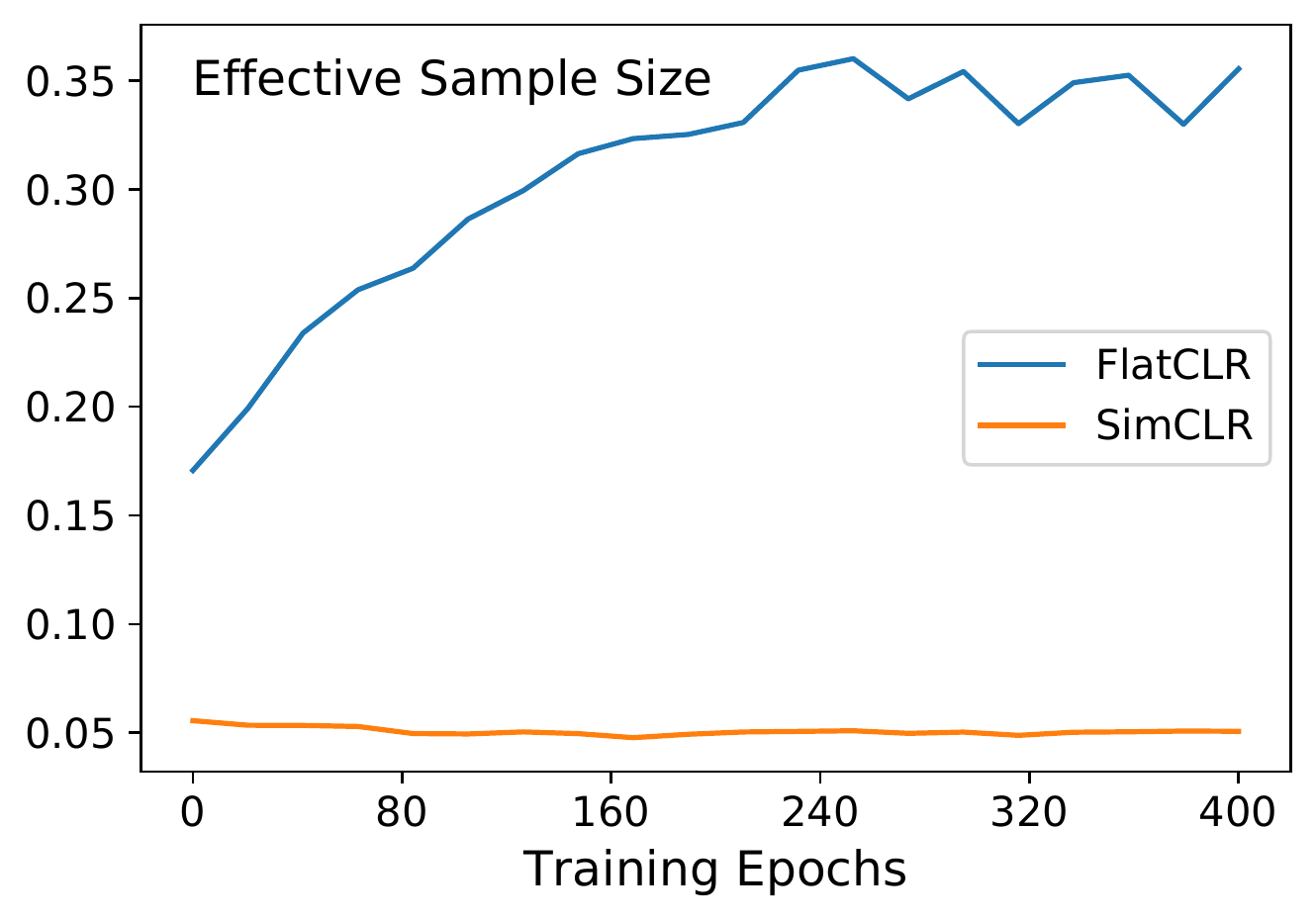}
  \vspace{-1.5em}
  \captionof{figure}{Effective sample size ({\it c.f.} Figure \ref{fig:mi_opt}). \label{fig:ess}}
%   \label{fig:test2}
\end{minipage}
 \hspace{2pt}
\begin{minipage}{.45\textwidth}
  \centering
  \includegraphics[width=1.\textwidth]{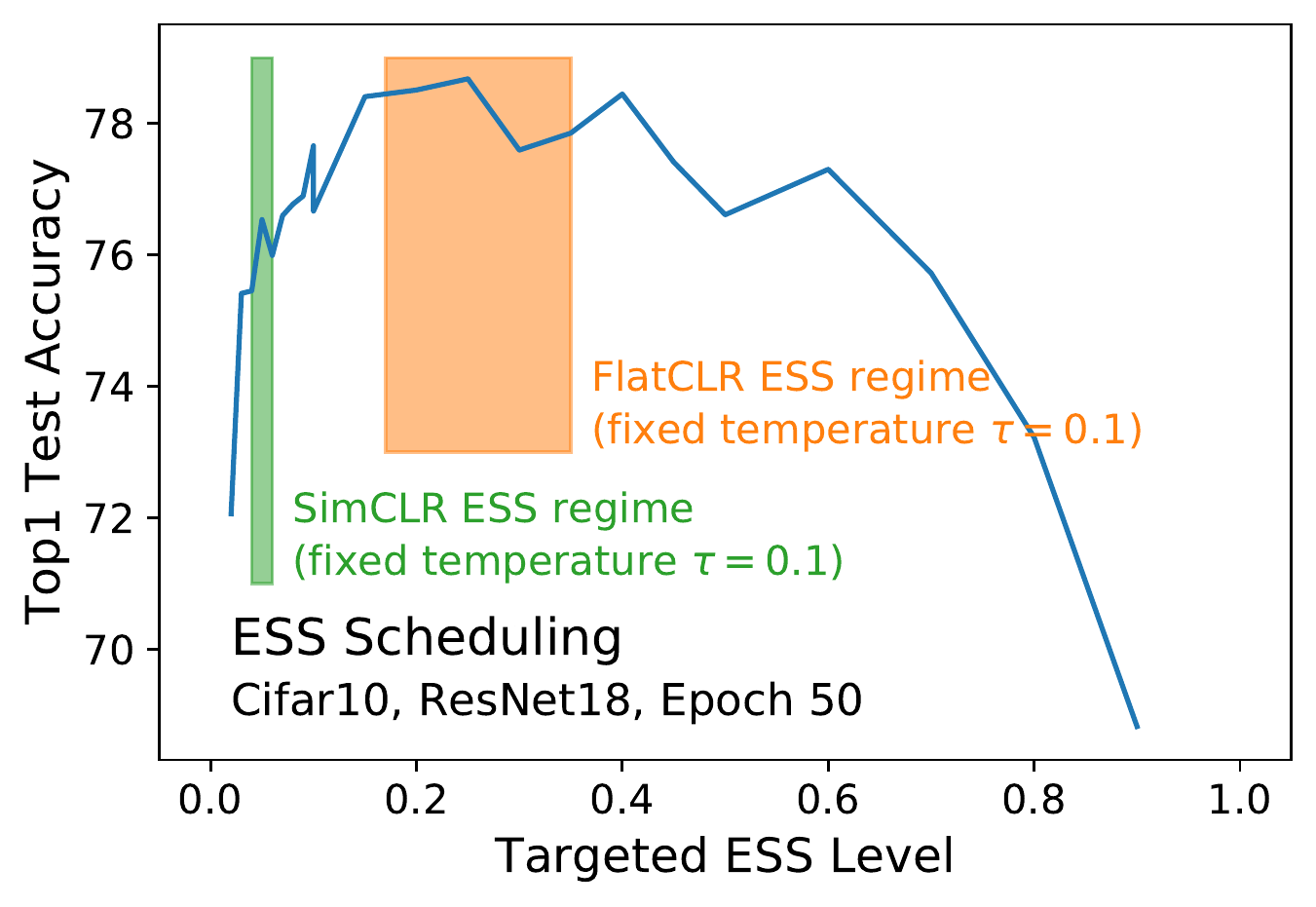}
  \vspace{-2.em}
  \captionof{figure}{ESS scheduling results. \label{fig:ess-acc}}
%   \label{fig:test2}
\end{minipage}
\vspace{-1.5em}
\end{figure}

% \begin{figure}[t!]
% \begin{center}
% %\includegraphics[width=.7\textwidth]{figures/results/cifar-flo-sim}
% \includegraphics[width=.48\textwidth]{figures/results/mi_acc}
% \end{center}
% \vspace{-1em}
% \caption{Self-MI of the learned representation strongly correlates with downstream performance. \label{fig:mi_acc}}
% \vspace{-1.em}
% \end{figure}

% \begin{figure}[t!]
% \begin{center}
% %\includegraphics[width=.7\textwidth]{figures/results/cifar-flo-sim}
% \includegraphics[width=.7\textwidth]{figures/results/ess}
% \end{center}
% \caption{Effective sample size during training. \label{fig:ess}}
% \end{figure}

{\bf Is tighter MI bound actually better or worse?} An interesting observation made by a few independent studies is that, perhaps contrary to expectation, tighter bounds on MI do not necessarily lead to better performance on the downstream tasks \citep{tschannen2020mutual}. To explain this, existing hypotheses have focused on the variance and sample complexity perspectives \citep{song2020understanding}. To address this, we compare the actual MI \footnote{Ground-truth MI is approximated by $\infonce$ using a very large negative sample pool.} to the mini-batch MI estimate, and plot the respective typical training curves in Figure \ref{fig:mi_opt}. Since $\FLAT$ itself is not associated with a number to bound MI (because it is theoretically tight), we use an $\infonce$ estimate based-on the $\FLAT$ representation. Observe that although the sample MI estimates are tied, $\FCLR$ robustly improves the ground-truth MI as $\SimCLR$ approaches the $\log$-$K$ saturation point and become stagnant. To further understand how MI relates to downstream performance, we plot the Top-$1$ accuracy against the true MI using all our model training checkpoints (Figure \ref{fig:mi_acc}), and confirm a strong linear relation between the two (Pearson correlation $\rho=0.65$, $p$-value $<10^{-20}$). However, this link is not evident using the mini-batch sample MI (Figure S1 in Appendix).

% An interesting observation from the literature is that, perhaps contrary to expectation, tighter bounds on MI not necessarily leads to better performance on the downstream tasks \citep{}. To explain this, existing theories have exclusively focused on the variance and sample complexity perspectives. 
{\bf ESS for monitoring and tuning contrastive learning.} As an important tool introduced in this work, we want to demonstrate the usefulness of $\ESS$ in contrastive training. Figure \ref{fig:ess} plots $\ESS$ curves for the training dynamics described in Figure \ref{fig:mi_opt}, and we see drastically different profiles. As predicted by our analyses, $\SimCLR$'s $\ESS$ monotonically decreases as it approaches the $\infonce$ saturation (from $0.06$ to $0.05$), while $\FLAT$-ESS instead climbs up ($0.17 \rightarrow 0.35$). The performance gap widens as the $\ESS$ difference becomes larger, thus confirming the superior sample efficiency of $\FLAT$. Next we experimented with $\ESS$-scheduling: instead of a fixed temperature, we fix the $\ESS$ throughout training, and then compare model performance. Figure \ref{fig:ess-acc} shows a snapshot of training progress per targeted $\ESS$ value at epoch $50$, where the estimated MI just started to plateau. The result indicates $\ESS$ range $[0.15, 0.4]$ works well for \texttt{Cifar10}, while $\SimCLR$ with fixed temperature only covers the sub-optimal $[0.04, 0.06]$. These interesting observation warrant further future investigations on $\ESS$ control in contrastive training. 
%$\FCLR$

{\bf Self-normalized contrastive learning as constrained optimization.} Here we want to promote a new view, which considers self-normalized contrastive learning as a form of constrained optimization. In this view, including multiple negative samples in the update of the critic function is necessary for contrastive learning. This conclusion comes from our numerous failed attempts in designing alternative few-sample contrastive learning objectives that simultaneously reduce estimation variance and tighten the MI bound (see Appendix for a detailed summary of our negative experience). Since the feature encoders are usually built with complex neural networks, the representations can be rather sensitive to the changes in encoder parameters. So while the gradient update direction may maximally benefit the MI estimate, it may disrupt the representation and thus compromise the validity of the variational MI estimate. Including negative samples in the updates of the critic $g_{\theta}$ allows the use of negative samples to provide instant feedback on which directions are bad, and to steer away from. More negative samples ({\it i.e.}, a larger $K$) will enforce a more confined search space, thus allowing the critic updates to proceed more confidently with larger learning rates. Also, comparison should be made to {\it importance-weighted variational auto-encoder} (IW-VAE) \citep{burda2015importance}, which also leverages a self-normalized objective for representation learning and inference. However, IW-VAE has been proven harmful to representation learning, although it provably tightens the likelihood bound \citep{rainforth2017tighter}. Finally, our new approach also promises to scale up \& improve {\it generalized contrastive learning} \citep{hyvarinen2019nonlinear}.

\begin{table}[t!]
\begin{center}
\caption{Comparison of representative variational MI objectives. We use $(x, p_{\oplus})$ to denote the positive sample drawn from the joint density $p(x,y)$, $(x, y_{\ominus})$ for the negative samples from $p(x)p(y)$, and $m(x, y^{1:K}) \triangleq \frac{1}{K} \sum_{k=1}^{K} \exp(g(x, y^k))$. See Appendix for more details. \label{tab:bounds}}
\vspace{-1.em}
\scalebox{.8}{
\begin{tabular}{cccccccc}
\toprule
Name & Objective & Bias & Stability &  \\
\midrule
{\it Donsker-Varadhan} \citep{donsker1983asymptotic} & $g(x, y_{\oplus}) - \log (\sum_{k=1}^{K}\exp(g(x^k, y_{\ominus}^k))/K)$ & Large & Poor \\
[4pt]
{\it Nguyen-Wainwright-Jordan} \citep{nguyen2010estimating} & $g(x,y_{\oplus}) - \sum_{k=1}^K \exp(g(x, y_{\ominus}^k)-1)/K$ & Low & Okay \\
[4pt]
{\it Fenchel-Legendre} \citep{guo2021tight}  & $u(x,y_{\oplus}) + \sum_{k=1}^K \exp(-u(x,y_{\oplus})+g(x,y_{\ominus})-g(x,y_{\oplus}))/K$ & Low & Okay \\
[4pt]
\texttt{InfoNCE} \citep{oord2018representation} & $g(x,y_{\oplus}) - \log(m(x, \{ y_{\oplus}, y_{\ominus}^{1:K-1}\}))$ & Large & Excellent \\
[6pt]
$\FLAT$ (Ours) & $\{m(x, y_{\ominus}^{1:K})-g(x,y_{\oplus})\}/\texttt{detach}[\{m(x,y_{\ominus}^{1:K})-g(x,y_{\oplus}\}]$ & Low & Excellent \\
\bottomrule
\end{tabular}
}
\vspace{-2.5em}
\end{center}
\end{table}

{\bf Connections to variational mutual information estimation.} Table \ref{tab:bounds} summarizes representative examples of nonparametric variational MI bounds in the literature, whose difference can be understood based on how information from negative samples are aggregated. Before $\infonce$, {\it Donsker-Varadhan} ($\DV$) \citep{donsker1983asymptotic} and {\it Nguyen-Wainwright-Jordan} ($\NWJ$) \citep{nguyen2010estimating} are the most widely practiced MI estimators. $\NWJ$ is generally considered non-contrastive as positive and negative samples are compared, respectively, at $\log$ and $\exp$ scales. $\DV$ differs from $\infonce$ by excluding the positive sample from the negative pool, which is similar to the practice of our $\FLAT$. However, $\DV$ is numerically unstable and necessitates careful treatment to be useful \citep{belghazi2018mutual}.
Also note some literature had unfairly compared the the multi-sample $\infonce$ to the single-sample versions of its competitors, partly because the alternatives do not have efficient multi-sample implementations. 
To the best of our knowledge, closest to this research is the concurrent work of \citep{guo2021tight}, where the contrastive {\it Fenchel-Lengendre} estimator is derived. While developed independently from completely different perspectives, $\FLAT$ enjoys the duality view promoted by \citep{guo2021tight} and inherits all its appealing theoretical properties. Our theoretical and empirical results complemented nicely the theories from \citep{guo2021tight}. 

\vspace{-8pt}
\section{Further Experiments}
\vspace{-5pt}
\label{sec:exp}

The above discussion presented several experimental results to highlight unique aspects of the proposed approach. We now consider additional experiments to further validate the proposed $\FLAT$ and benchmark it against state-of-the-art solutions. We sketch our setup here and leave details to the Appendix. Our code can be assessed from \url{https://github.com/Junya-Chen/FlatCLR}. All experiments are implemented with \texttt{PyTorch} and executed on NVIDIA V100 GPUs with a maximal level of parallelism at 4 GPUs. 

% Details of the experimental setup are described the Appendix, and 

{\bf Self-supervised learning (SSL) on Cifar and ImageNet.} We set our main theme in SSL and compare the effectiveness of the $\SimCLR$ framework \citep{chen2020simple} to our $\FLAT$-powered $\FCLR$. Our codebase is modified from a public \texttt{PyTorch} implementation\footnote{\url{https://github.com/sthalles/SimCLR}}. Specifically, we train $256$-dimensional feature representations by maximizing the self-MI between two random views of data, and report the test set classification accuracy using a linear classifier trained to convergence. We report performance based on \texttt{ResNet-50}, and some of the learning dynamics analyses are based on \texttt{ResNet-18} for reasons of memory constraints. Hyper-parameters are adapted from the original $\SimCLR$ paper. For the large-batch scaling experiment, we first grid-search the best learning rate for the base batch-size, then grow the learning rate linearly with batch-size. 

\begin{table}[t!]
\caption{ImageNet SSL results. \label{tab:imagenet_ssl}}
\scalebox{1.}{
\setlength{\tabcolsep}{5pt}
\begin{tabular}{cccccccccccc}
\toprule
Epoch   & 10    & 20    & 30    & 40    & 50    & 60    & 70    & 80    & 90    & 100   \\
\midrule
$\SimCLR$  & 38.57 & 43.71 & 47.03 & 49.45 & 49.93 & 52.18 & 53.31 & 53.47 & 53.98 & 54.62 \\ 
$\FCLR$  & 39.7  & 45.30  & 47.74 & 49.72 & 50.39 & 53.30  & 54.48 & 54.43 & 56.25 & 56.74 \\
\bottomrule
\end{tabular}
}
\end{table}

% We compare the effectiveness of self-supervised representation learning in the $\SimCLR$ framework. Our codebase is modified from a public \texttt{PyTorch} implementation \footnote{\url{https://github.com/sthalles/SimCLR}}. We use the \texttt{Cifar} dataset \citep{krizhevsky2009learning} for demonstration. We use the default setup from the codebase: a \texttt{ResNet-50} is used for feature extraction, where the feature dimension is set to $d=256$, temperature to $\tau=0.07$, learning rate to $10^{-4}$, with input images augmented at random. We used a model pre-trained to $60\%$ accuracy to hot start the training. A three-layer is fitted MLP after each training epoch using the learned representation to predict the labels. The \texttt{Adam} optimizer is used to train our model \citep{kingma2014adam}. We summarize the comparison of $\infonce$ and $\FLAT$ in Figure \ref{fig:batchsize}. 

The observations made on \texttt{Cifar} align with our theoretical prediction (see Figure \ref{fig:batchsize}): in the early training (less than $50$ epochs), where the contrast between positives and negatives have not saturated, all models performed similarly. After that, performance start to diverge when entering a regime where $\FLAT$ learns more efficiently. See Section \ref{sec:discuss} and Appendix for more results and discussions.

% SimCLR
% Cifar10:train_top1 83.37628936767578 test_top1 87.74671173095703 test_top5 99.73273468017578(1e-3)
% Cifar100:train_top1 77.67799377441406 test_top1 65.39884948730469 test_top5 90.96422576904297
% VOC: train_top1 77.86458587646484 test_top1 69.37934112548828 test_top5 95.94184112548828(1e-4)
% Flower:train_top1 100.0 test_top1 90.0390625 test_top5 98.2421875
% Caltech:train_top1 100.0 test_top1 79.541015625 test_top5 96.37044525146484(1e-3)
% SUN:train_top1 99.98458099365234 test_top1 49.624794006347656 test_top5 78.75719451904297(1e-4)

% Flat
% - Cifar10:train_top1 83.48098754882812 test_top1 87.92146301269531 test_top5 99.70188903808594
% - Cifar100:train_top1 66.45457458496094 test_top1 65.75863647460938 test_top5 90.5838851928711
% - VOC:train_top1 81.53211975097656 test_top1 69.66146087646484 test_top5 96.02864837646484
% - Flower:train_top1 100.0 test_top1 90.234375 test_top5 97.4609375 (1e-3)
% - Caltech:train_top1 98.671875 test_top1 81.23372650146484 test_top5 96.875 (1e-4>1e-3)
% - SUN:train_top1 94.80365753173828 test_top1 51.31578826904297 test_top5 80.65892028808594(1e-4)

\begin{table}[t!]
    \caption{ImageNet SSL transfer learning results.}
    \label{tab:imagenet_trans}
    \centering
    \begin{tabular}{ccccccc}
    \toprule
         Dataset & Cifar10 & Cifar100 & VOC2007 & Flower & Caltech101 & SUN397\\
        \midrule
         $\SimCLR$ & 87.74 & 65.40 & 69.38 & 90.03 & 79.54 & 49.62 \\
         $\FCLR$ & 87.92 & 65.76 & 69.66 & 90.23 & 81.23 & 51.31\\
    \bottomrule
    \end{tabular}
    \vspace{-1em}
\end{table}

We further apply our model to the \texttt{ImageNet} dataset and compare its performance to the $\SimCLR$ baseline. We note the SOTA results reported by \citep{chen2020simple} heavily rely on intensive automated hyper-parameter grid search, and considerably larger networks ({\it i.e.}, \texttt{ResNet50} $\times 4$ versus \texttt{ResNet50}), that we are unable to match given our (university-based) computational resources. So instead, we report fair comparison to the best of our affordability. Table \ref{tab:imagenet_ssl} reports SSL classification performance comparison up to the $100$ epoch\footnote{The reported results is a lower bound to actual performance. We were able to considerably improve the final result via running longer linear evaluation training with larger batch-sizes.}. In Table \ref{tab:imagenet_trans} we examine the performance of representation transfer to other datasets. For both cases, $\FCLR$ consistently outperforms the vanilla $\SimCLR$.

\vspace{-8pt}
\section{Conclusions}
\vspace{-5pt}

We have presented a novel contrastive learning objective called $\FLAT$, that is easy to implement, but delivers strong performance and faster model training. We show that underneath its simple expression, $\FLAT$ has a solid mathematical grounding, and consistently outperforms its $\infonce$ counterpart for the experimental setting we considered. In future work, we seek to verify the effectiveness of $\FLAT$ on a computation scale not feasible to this study, and apply it to new architectures and applications. Also, we invite the community to find ways to reconcile the performance gap between those theoretically optimal MI bounds and those self-normalized sub-optimal bounds such as $\FLAT$ and $\infonce$, and develop principled theories for hard-negative training. 

\bibliography{fclr}
\bibliographystyle{ieeetr}

\newpage

\appendix

\renewcommand{\thetable}{S\arabic{table}}
\renewcommand{\thefigure}{S\arabic{figure}}
\renewcommand{\thealgorithm}{S\arabic{algorithm}}
\renewcommand{\thethm}{S\arabic{thm}}

\setcounter{table}{0}
\setcounter{figure}{0}
\setcounter{algorithm}{0}
\setcounter{thm}{0}

%%%%% To create table of contents only for Appendix %%%%%%

\addcontentsline{toc}{section}{Appendix} % Add the appendix text to the document TOC
\part{Appendix} % Start the appendix part
\parttoc % Insert the appendix TOC
%%%%%%%%%%%%%%%%%%%%%%%%%%%%%%%%%%

\section{The Staggering Cost of Training Contrastive Learners}

In Table \ref{tab:cost} we summarize the associated cost of training state-of-the-art contrastive learners. We have used the numbers from the original papers to compute the cost. The number of devices and time of training for the largest model reported in the respective papers are used, while we use the online quotes from Google Cloud (for TPU units) and Amazon AWS (for GPU units) for the hourly cost of dedicated computing devices. We only focused on the computation cost, so the potential charges from storage and network traffic are omitted. Note that this table only reports the number of computing devices used in the final training where all parameters have been tuned to optimal, the actual expenditures associated with the development of these models can be significantly higher. Usually researchers and engineers spent more time tuning the parameters and exploring ideas before finally come up with a model that can be publicized. Also, the cost for performance evaluation is not count towards the cost, and some of the papers have employed intensive grid-search of parameters for evaluation, which in our experience can be even more costly than training the contrastive learners at times. And we do find fine-tuning evaluation can drastically boost the performance metrics.

\begin{table}[H]
\begin{center}
\caption{Cost of training a contrastive learner \label{tab:cost}}
\vspace{-.5em}
\begin{tabular}{cccccc}
\toprule
Model & Sponsor & Neg. Size & Infrastructure & Train Time & Est. Cost\\
\midrule
\texttt{MoCo} \citep{he2020momentum} & Facebook & $65,536$ & $64$ V100 GPUs & $6$ days & \$23k \\
\texttt{SimCLR} \citep{chen2020simple} & Google &  $4,096$ & $128$-core TPU-v3 & $15$ hours & \$1,720\\
\texttt{CLIP} \citep{radford2021learning} & OpenAI & $32,768$ & $592$ V100 GPUs & $18$ days & \$634k \\
\bottomrule
\end{tabular}
\end{center}
\vspace{-2.em}
\end{table}

\section{Technical Proofs}

\subsection{Proof of Proposition 2.1}

\begin{proof}
See \citep{poole2019variational} for a neat proof on how the multi-sample $\NWJ$ upper bounds $\infonce$. Since $\NWJ$ is a lower bound to MI, $\infonce$ also lower bounds MI. 
% So for arbitrary multi-sample critic $f(x; y_{1:K})$, we know 
% \beq
% I(X;Y) = I(X_1;Y_{1:K}) \geq I_{\NWJ}(X_1,Y_{1:K};f) = \EE_{p(x_1,y_1)\prod_{k>1}p(y_k)}[f(x_1,y_{1:K})]-e^{-1}\EE_{p(x)}[Z_f(x)]
% \eeq

% Now let us set 
% \beq
% \tilde{f}(x_1;y_{1:K}) = 1 + \log \frac{e^{g(x_1, y_1)}}{m(x_1;y_{1:K})}, \quad m(x_1;y_{1:K}) = \frac{1}{K} \sum_k e^{g(x_1,y_k)}. 
% \eeq

% \beqs
% I_{\NWJ}(X_1,Y_{1:K};\tilde{f}) & = & \EE_{p(x_1,y_1)p^{K-1}(y_k)}\left[ 1+ \log \frac{e^{g(x_1, y_1)}}{m(x_1;y_{1:K})} \right] - \EE_{p(x')p^K(y')} \left[e^{-1+1 +\log \frac{e^{g(x_1', y_1')}}{m(x_1'; y_{1:K}' )} }\right] \\
% & = & \EE_{p(x_1,y_1)p^{K-1}(y_k)}\left[ 1+ \log \frac{e^{g(x_1, y_1)}}{m(x_1;y_{1:K})} \right] - \EE_{p(x')p^K(y')} \left[ \frac{e^{g(x_1', y_1')}}{m(x_1'; y_{1:K}' )} \right] 
% \eeqs
% Due to the symmetry of $\{y_k\}_{k=1}^K$, we have 
% \beq
% \EE_{p(x')p^K(y')} \left[ \frac{e^{g(x_1', y_1')}}{m(x_1'; y_{1:K}' )} \right]  = \EE_{p(x')p^K(y')} \left[ \frac{e^{g(x_1', y_k')}}{m(x_1'; y_{1:K}' )} \right].
% \eeq
% So this gives
% \beq
% \EE_{p(x')p^K(y')} \left[ \frac{e^{g(x_1', y_1')}}{m(x_1'; y_{1:K}' )} \right] = \EE_{p(x')p^K(y')} \left[ \frac{\frac{1}{K}e^{g(x_1', y_k')}}{m(x_1'; y_{1:K}' )} \right]  = 1,
% \eeq
% which proves
% \beq
% I_{\NWJ}(X_1,Y_{1:K};\tilde{f}) = \EE_{p(x_1,y_1)p^{K-1}(y_k)}\left[\log \frac{e^{g(x_1, y_1)}}{m(x_1;y_{1:K})} \right] \triangleq I_{\infonce}
% \eeq

What remains is to show the $\infonce$ bound is asymptotically tight. We only need to prove that with a specific choice of $f(x,y)$, $\infonce$ recovers $I(X;Y)$. To this end, let us set $f(x,y) = f^*(x,y) = \frac{p(y|x)}{p(y)}$, and we have
\beqs
I_{\infonce}^K(f^*) & = & \EE_{p^K} \left[ \log \left( \frac{f^*(x_k,y_k)}{f^*(x_k,y_k) + \sum_{k'\neq k} f^*(x_k,y_{k'})} \right) \right] + \log K \\
& = & -\EE\left[ \log \left( 1+\frac{p(y)}{p(y|x)} \sum_{k'} \frac{p(y_{k'}|x_k)}{p(y_{k'})} \right)\right] + \log K \\
& \approx & -\EE\left[ \log \left( 1+\frac{p(y)}{p(y|x)} (K-1) \EE_{y_{k'}} \frac{p(y_{k'}|x_k)}{p(y_{k'})} \right)\right] + \log K\\
& = &  -\EE\left[ \log \left( 1+\frac{p(y_k)}{p(y_k|x_k)} (K-1) \right)\right]+\log K\\
& \approx & \underbrace{-\EE\left[\log\frac{p(y)}{p(y|x)}\right]}_{I(X;Y)} - \log (K-1) + \log K\\
% (K\rightarrow \infty) & \rightarrow & I(X;Y)
\eeqs
Now taking $K\rightarrow \infty$, the last two terms cancels out.
\end{proof}

\subsection{Proof of Proposition 3.1}

\begin{proof}

Without loss of generality we denote $y_0$ as the positive sample and all $y_j, j>0$ as the negative samples. Recall 
\beqs
&& \texttt{CrossEntropyLoss}(\texttt{logits} = g_{\theta}(x_0,y_j), \texttt{label}=0) \\
&=& -\log \frac{\exp(g_{\theta}(x_0,y_0))}{\sum_j \exp(g_{\theta}(x, y_j))} \\
&=& \log \sum\nolimits_j \exp(g_{\theta}(x_0, y_j)-g_{\theta}(x_0, y_0))
\eeqs
Since $\nabla \log f = \frac{\nabla f}{f}$, so 
\beq
\nabla_{\theta} I_{\FLAT}^{\oplus}(g_{\theta}) = \nabla_{\theta} \CL_{\texttt{CrossEntropy}} = \frac{\nabla_{\theta} \{\sum\nolimits_j \exp(g_{\theta}(x_0, y_j)-g_{\theta}(x_0, y_0))\}}{\sum\nolimits_j \exp(g_{\theta}(x_0, y_j)-g_{\theta}(x_0, y_0))} = \nabla_{\theta} I_{\infonce}(g_{\theta})
\label{eq:grad}
\eeq
which concludes our proof (we omit the sign here for brevity).
\end{proof}

\subsection{Proof of Proposition 3.2}

\begin{proof}
Let us pick up from (\ref{eq:grad}) from last proof, we have
\beqs
\nabla I_{\FLAT} & = & \frac{\nabla_{\theta} \{\sum\nolimits_j \exp(g_{\theta}(x_0, y_j)-g_{\theta}(x_0, y_0))\}}{\sum\nolimits_j \exp(g_{\theta}(x_0, y_j)-g_{\theta}(x_0, y_0))}\\
& = & \frac{\sum\nolimits_j \exp(g_{\theta}(x_0, y_j)-g_{\theta}(x_0, y_0))(\nabla_{\theta} \{ g_{\theta}(x_0, y_j)-g_{\theta}(x_0, y_0) \})}{\sum\nolimits_j \exp(g_{\theta}(x_0, y_j)-g_{\theta}(x_0, y_0))} \\
& = & \sum\nolimits_j w_j \nabla_{\theta} g_{\theta}(x_0, y_j) - (\sum\nolimits_j w_j) \nabla_{\theta} g_{\theta}(x_0, y_0) \\
& = & \sum\nolimits_j w_j \nabla_{\theta} g_{\theta}(x_0, y_j) - g_{\theta}(x_0, y_0)
\eeqs
here $w_j \triangleq \exp(g_{\theta}(x_0, y_j)) / (\sum\nolimits_{j'} \exp(g_{\theta}(x_0, y_{j'})))$, 
as the term $\exp(-g_{\theta}(x_0,y_0))$ has been canceled out. 
\end{proof}

\subsection{Proof of Lemma 3.3}

% Our proof follows some of the theoretical results developed in \citep{guo2021tight}, which are included here for completeness. 

Our proof is inspired by the technique used in \citep{tao2019fenchel} for non-parametric likelihood approximations, which is based on the celebrated Fenchel-Legendre duality given below.

\begin{defn}[Fenchel-Legendre duality \citep{fenchel1949conjugate}]
Let $f(t)$ be a proper convex, lower-semicontinuous function; then its {convex conjugate} function $f^*(v)$ is defined as 
$f^*(v) = \sup_{t\in \CD(f)}\{ t v - f(t) \}$, where $\CD(f)$ denotes the domain of function $f$ \citep{hiriart2012fundamentals}. We call $f^*(v)$ the \textit{Fenchel-Legendre conjugate} of $f(t)$, which is again convex and lower-semicontinuous. The Fenchel-Legendre conjugate pair $(f, f^*)$ are dual to each other, in the sense that $f^{**} = f$, \textit{i.e.}, 
$
f(t) = \sup_{v\in \CD(f^*)}\{ v t - f^*(v) \}. 
$
\end{defn}

{\bf Example.} The Fenchel-Legendre dual for $f(t) = -\log(t)$ is $f^*(v) = -1-\log(-v)$. 

% {\it Proof of Lemma 3.3.}
% \begin{proof}
{\it Proof of Lemma 3.3.} \\
Let us write $\infonce$ as 
\beq
I_{\infonce}(g) = -\log \sum\nolimits_j \exp(g_{\theta}(x_0, y_j)-g_{\theta}(x_0, y_0)).
\eeq
Replacing the $-\log(t)$ term in $I_{\infonce}(t)$ with its Fenchel-Legendre dual $-1-\log(-v)$, then Proposition \label{thm:fince} is immediate after properly rearranging the terms and write $u=-\log v$. \hfill $\square$

\subsection{Proof of Corollary 3.4}

To make our proof simpler, we follow some theoretical results developed in \citep{guo2021tight}, included below for completeness.

\begin{prop}[The Fenchel-Legendre Optimization Bound, Proposition 2.2 in \citep{guo2021tight}]
\label{thm:fince}
\beqs
& I_\FLO(u, g) \triangleq \left\{\EE_{p(x,y)p(y')}\left[u(X,Y) +\exp(-u(X,Y)+g(X,Y')-g(X,Y))\right]\right\}+1 \\
[3pt]
& I(X;Y) = - \min_{u,g} \{I_\FLO(u, g)\} \label{eq:fnce}
\eeqs
\end{prop}

{\it Sketch of proof for Proposition \ref{thm:fince}.} Recall the 
{\it Donsker-Varadhan} (DV) bound \citep{donsker1983asymptotic} is given by 
\beq
I_{\DV} \triangleq \EE_{p(x,y)}[g(x, y) - \log (\EE_{p(y')}[\exp(g(x, y'))])].
\eeq
Then we proceed similarly to the proof of Lemma 3.3. 

{\it Remark.} Here we consider $g(x,y)$ as the primal critic and $u(x,y)$ as the dual critic. Since arbitrary choice of primal/dual critics always lower bounds MI, we can either jointly optimize the two critics, or train in an iterative fashion: optimize one at a time while keep the other fixed. Let us consider the case $u$ is fixed and only update $g$, the proof below shows with an appropriate choice of $u$, Corollary 3.4 follows.

{\it Proof of Corollary 3.4}\\
Given $g_{\theta}(x,y)$ and empirical samples $\{ (x_j,y_j) \}$, let us set $u(x,y)$ to 
\beq
\hat{u}^*(g_{\theta}) = \log \left(\frac{1}{K}\sum\nolimits_{j} \exp(g_{\theta}(x_i, y_j) - g_{\theta}(x_i, y_i)) \right)
\eeq
Plug $(g_{\theta}, \hat{u}^*)$ into the right hand side of Equation (9) proves $\hat{u}^* + I_{\FLAT} - 1$ lower bounds mutual information. Since $\hat{u}^*$ does not contribute gradient, we can consider $I_{\FLAT} \leq I(X;Y)$ holds up to a constant term. In other words, we are effectively optimizing a lower bound to MI, although $I_{\FLAT}$ does not technically a lower bound --  this is still OK since the difference does not contribute learning signal.
\hfill $\square$

% \beq
% \mathfrak{u}_{\theta}(\{ (x_i, y_i) \}) = \log \left(\frac{1}{K}\sum\nolimits_{j} \exp(g_{\theta}(x_i, y_j) - g_{\theta}(x_i, y_i)) \right)
% \label{eq:ug}
% %_{i=1}^K
% \eeq 
% and update $u_{\theta}(x,y)$ while artificially keeping $g_{\theta}(x,y)$ fixed \footnote{That is to say $g_{\theta}$ in $u_{\phi}$ is an independent copy of $g_{\theta}$.}, then $\FLO$ falls back to $\DV$. Alternatively, we can consider the Fenchel dual version of it: using the same multi-input $u_{\phi}(\{x_i, y_i \})$ in (\ref{eq:ug}), treat $u_{\phi}$ as fixed and only update $g_{\theta}$, and this gives us the following novel MI objective we call {\it Fenchel-Donsker-Varadhan} (FDV) estimator: 
% \beq
% I_{\FDV} \triangleq \hat{I}_{\DV}(\{ (x_i, y_i) \}) + \frac{\sum_{j} \exp(g_{\theta}(x_i, y_j) - g_{\theta}(x_i, y_i))}{\sum_{j} \exp(\hat{g}_{\theta}(x_i, y_j) - \hat{g}_{\theta}(x_i, y_i))} - 1, 
% \eeq
% where we have used $\hat{g}, \hat{I}$ to denote evaluation-only mode for the corresponding functions (because they are the ``fixed'' $u_{\phi}$ and do not backpropagate parameter gradients). 

% \begin{col}[Dual estimation of MI, \citep{guo2021tight} Corollary 2.3]
% \label{thm:fince_mi}
% Let $u^*(x,y)$ be the solution for (\ref{eq:fnce}), then we have
% \beq
% I(X,Y) = \EE_{p(x,y)}[-u^*(X,Y)]. 
% \eeq
% \end{col}

\subsection{Proof of Proposition 3.7}

\begin{proof}
% m_\gamma(\{ a_i \}_{i=1}^n) = \left(\frac{1}{n}\sum_i a_i^\gamma \right)^{\frac{1}{\gamma}}
Denoting $f_j = \exp(g_j)$, and we have
\beqs
\nabla I_{\gamma}(g_{\theta}) & = & \frac{\nabla m_{\gamma}(\{f_j\})}{m_{\gamma}(\{f_j\})} \\
& = & \frac{\frac{1}{\gamma}(\frac{1}{n}\sum_j f_j^{\gamma})^{\frac{1}{\gamma}-1}\{ \gamma \frac{1}{n}\sum_j f_j^{\gamma-1} \nabla f_j \}}{(\frac{1}{n}\sum_j f_j^{\gamma})^{\frac{1}{\gamma}}} \\
& = & \frac{\sum_j f_j^{\gamma-1} \nabla f_j}{\sum_j f_j^{\gamma}} \\
& = & \frac{\nabla \sum_j \exp(\gamma g_j)}{\gamma\sum_j \exp(\gamma g_j)} \\
& = & \frac{1}{\gamma}\nabla I_{\FLAT}(\gamma \cdot g_{\theta})
% \frac{m_{\gamma-1}(\{g_j\}) \nabla \{\sum_j g_j^{\gamma} \}}{m_{\gamma}(\{g_j\})}
\eeqs
\end{proof}

\subsection{Proof of Proposition 3.8}

Here we detail the technical conditions for Proposition 3.8 to hold. Our derivation follows the analytic framework of generalized SGD from \citep{tao2019fenchel}, included below for completeness. 

\begin{defn}[Generalized SGD, Problem 2.1 \citep{tao2019fenchel}] 
Let $h(\theta;\omega), \omega \sim p(\omega)$ be an unbiased stochastic gradient estimator for objective $f(\theta)$, $\{ \eta_t > 0 \}$ is the fixed learning rate schedule, $\{\xi_t>0\}$ is the random perturbations to the learning rate. We want to solve for $\nabla f(\theta) = 0$ with the  iterative scheme
$
\theta_{t+1} = \theta_t + \tilde{\eta}_t \, h(\theta_t; \omega_t), 
$
where $\{\omega_t\}$ are iid draws and $\tilde{\eta}_t = \eta_t \xi_t$ is the randomized learning rate. 
\end{defn}

\begin{assumption}(Standard regularity conditions for SGD, Assumption D.1 \citep{tao2019fenchel}).
\label{thm:assum}
\vspace{-1em}
\begin{enumerate}
\setlength\itemsep{2pt}
\item[$A1.$] $h(\theta)\triangleq\EE_{\omega}[h(\theta;\omega)]$ is Lipschitz continuous;
\item[$A2.$] The ODE $\dot{\theta} = h(\theta)$ has a unique equilibrium point $\theta^*$, which is globally asymptotically stable;
\item[$A3.$] The sequence $\{ \theta_t \}$ is bounded with probability one;
\item[$A4.$] The noise sequence $\{ \omega_t \}$ is a martingale difference sequence;
\item[$A5.$] For some finite constants $A$ and $B$ and some norm  $\| \cdot \|$ on $\BR^d$, $\EE[\| \omega_t \|^2] \leq A + B \| \theta_t \|^2$ almost surely $\forall t \geq 1$. 
\end{enumerate}
\end{assumption}
% {\it Remark.} In the context of stochastic optimization, the globally asymptotic stability can be implied, for example, when $f(\theta)$ is strict convex (recall $h(\theta) = \nabla f(\theta)$). 

\begin{prop}[Generalized SGD, Proposition 2.2 in \citep{tao2019fenchel}] 
\label{thm:gsa}
Under the standard regularity conditions listed in Assumption \ref{thm:assum}, we further assume $\sum_t \EE[\tilde{\eta}_t] = \infty$ and $\sum_t \EE[\tilde{\eta}_t^2] < \infty$. 
Then $\theta_n \rightarrow \theta^*$ with probability one from any initial point $\theta_0$. 
\end{prop}

 \begin{assumption}(Weak regularity conditions for generalized SGD, Assumption G.1 in \citep{tao2019fenchel}).
\label{thm:assum_gen}
\vspace{-1em}
\begin{enumerate}
\setlength\itemsep{2pt}
	\item[$B1.$] The objective function $f(\theta)$ is second-order differentiable;
	\item[$B2.$] The objective function $f(\theta)$ has a Lipschitz-continuous gradient, i.e., there exists a constant $L$ satisfying 
		$-LI\preceq \nabla^2f(\theta)\preceq LI$, where for semi-positive definite matrices $A$ and $B$, $A \preceq B$ means $v^T A v \leq v^T B v$ for any $v\in \BR^d$;
	\item[$B3.$] The noise has a bounded variance, i.e., there exists a constant $\sigma>0$ satisfying $\mathbb{E}\left[\left\|h(\theta_t;\omega_t) - \nabla f(\theta_t)\right\|^2\right]\leq \sigma^2$.
\end{enumerate}
\end{assumption}

\begin{prop}[Weak convergence, Proposition G.2 in \citep{tao2019fenchel}]
\label{thm:weak_conv}
Under the technical conditions listed in Assumption \ref{thm:assum_gen}, the SGD solution $\{ \theta_t \}_{t>0}$ updated with generalized Robbins-Monro sequence ($\tilde{\eta}_t$: $\sum_t \EE[\tilde{\eta}_t] = \infty$ and $\sum_t \EE[\tilde{\eta}_t^2] < \infty$) converges to a stationary point of $f(\theta)$ with probability $1$ (equivalently, $\mathbb{E}\left[\|\nabla f(\theta_t)\|^2\right]\rightarrow 0$ as $t\rightarrow \infty$).
\end{prop}

{\it Proof of Proposition 3.8.}\\
For fixed $g_{\theta}(x,y)$ the corresponding optimal $u_{\theta}^*(x,y)$ maximizing the rhs in Equation (9) is given by 
\beq
u_{\theta}^*(x, y) = \log \EE_{p(y')}[ \exp(g_{\theta}(x,y') - g_{\theta}(x,y))] \triangleq -\log \CE_{\theta}(x,y), 
\eeq
so $\hat{\CE}_{\theta}(x,y) \triangleq \exp^{-\hat{u}_{\phi}(x,y)}$ can be considered as approximations to $\CE_{\theta}(x,y)$. 
% this relation implies the term $\exp^{-u_{\phi}(x,y)}$ is essentially optimized to approximate $\CE_{\theta}(x,y)$. To emphasize this point, we now write $\hat{\CE}_{\theta}(x,y) \triangleq e^{-u_{\phi}(x,y)}$. 
% When this approximation is sufficiently accurate ({\it i.e.}, $\CE_{\theta} \approx \hat{\CE}_{\theta}$), we can show $\nabla I_{\FLO}$ approximates $\nabla I_{\UBA}$ as follows
\beqs
\nabla_{\theta}\{ (9) \} & = & - \EE_{p(x,y)}\left[ e^{-u_{\phi}(x,y)}\EE_{p(y')}[ \nabla_{\theta} \exp(g_{\theta}(x,y') - g_{\theta}(x,y))]\right] \\
%& = &  \EE_{p(x,y)}\left[ \frac{e^{-u_{\phi}(x,y)}}{\CE_{\theta}(x,y)} \nabla_{\theta} \log \CE_{\theta}(x,y)\right]\\
& = &  \EE_{p(x,y)}\left[ \frac{\hat{\CE}_{\theta}(x,y)}{\CE_{\theta}(x,y)} \nabla_{\theta} \log \CE_{\theta}(x,y)\right] %\approx \EE_{p(x,y)}\left[  \nabla_{\theta} \log \CE_{\theta}(x,y) \right] \label{eq:grad_scale}\\
% & = &  \nabla_{\theta} \left\{ \EE_{p(x,y)}[\log \CE_{\theta}(x,y)] \right\} = \nabla_{\theta} \{I_{\UBA}(g_\theta)\}.
\eeqs
Note $I_{\BA} \triangleq \max_{g_{\theta}}\{\EE_{p(x,y)}[\log \CE_{\theta}(x,y)]\}$ is the well-known {\it Barber-Agakov} (BA) representation of mutual information ({\it i.e.}, $I_{\BA} = I(X;Y)$) \citep{agakov2004algorithm, poole2019variational}, so optimizing Equation (9)\footnote{Based on the proof of Corollary 3.4, we know $\FLAT$ optimization is a special case of optimizing Equation (9). } with SGD is equivalent to optimize $I_{\BA}$ with its gradient scaled (randomly) by $\hat{\CE}_{\theta_t}/\CE_{\theta_t}$ \citep{guo2021tight}.
% Indeed, we can prove $\FLO$ converges under much weaker conditions, even when this approximation $\hat{u}(x,y)$ is very rough. 
% The intuition is simple: in (\ref{eq:grad_scale}), the term $\frac{\hat{\CE}_{\theta_t}}{\CE_{\theta_t}}$  only rescales the gradient, which implies the optimizer will still proceed in the right direction. 
Under the additional assumption that $\hat{\CE}_{\theta_t}/\CE_{\theta_t}$ is bounded between $[a,b]$ ($0<a<b<\infty$), results follow by a direct application of Proposition \ref{thm:gsa} and Proposition \ref{thm:weak_conv}. \hfill $\square$

\section{Algorithm for ESS Scheduling}

We summarize the effective-sample size ($\ESS$) scheduling scheme in Algorithm \ref{alg:ess_anneal}.

\begin{algorithm}[H]
\caption{$\ESS$ Scheduling}
   \label{alg:ess_anneal}
\begin{algorithmic}
%\small
\STATE Empirical data distribution $\hat{p}_d = \{ (x_i, y_i) \}_{i=1}^n$ \\
[1pt]
\STATE Inverse temperature $\beta=1$, ESS-scheduler $\{\varrho_t\in (1/K, 1]\}_{t=1}^T$\\
[1pt]
\STATE Adaptation rate $\gamma = 0.01$\\
[3pt]
\FOR{$t=1,2,\cdots, T$}
\STATE Sample $i,i_k' \sim [1, \cdots, n], k'\in[1, \cdots, K]$ \\
[1pt]
\STATE $\gv_{\oplus} = g_{\theta}(x_i, y_i), \gv_{\ominus} = g_{\theta}(x_i, y_{i_k'})$\\
[1pt]
\STATE $\texttt{clogits} = \texttt{logsumexp}(\gv_{\ominus} - \gv_{\oplus})$\\
[1pt]
\STATE $\texttt{weights} = \texttt{Softmax}(\gv_{\ominus} - \gv_{\oplus})$\\
[1pt]
\STATE $\texttt{ESS} = 1./(K \cdot \texttt{square}(weights).sum())$\\
% \STATE $\beta = \?0.99\beta:1.01 \beta$
\STATE $\ell_{\FLAT} = \exp(\texttt{clogits}-\texttt{detach}[\texttt{clogits}])$ \\
[1pt]
\STATE \# Use your favorite optimizer
\IF{$\texttt{ESS}>\varrho_t$} \STATE {$\beta = (1-\gamma) \cdot \beta$} \ELSE \STATE{$\beta = (1+\gamma) \cdot \beta $} \ENDIF
\ENDFOR
\end{algorithmic}
\end{algorithm}

\section{Failed Attempts to Overcome the $\log$-K Curse}

The author(s) feel it is imperative to share not only successful stories, but more importantly, those failure experience when exploring new ideas. We contribute this section in the hope it will both help investigators avoid potential pitfalls and inspire new researches. 

{\bf Joint optimization of primal-dual critics.} Inspired by the concurrent research of \citep{guo2021tight}, the author(s) of this paper had originally hope the joint optimization of primal-dual critic as defined in Equation (9) will match, and hopefully surpass the performance of multi-sample $\infonce$ with single-sample estimation ({\it i.e.}, $K=1$). The argument is follows: in theory, the single-sample Fenchel-Legendre estimator has the same expectation with its multi-sample variant, and is provably tighter than $\infonce$. In a sense, Fenchel-Legendre estimator is combining the gradient of $\FLAT$ and $\infonce$, and the potential synergy is appealing. Unfortunately, in our small scale trial experiments ({\it i.e.}, MNIST and Cifar), we observe that while the Fenchel-Legendre estimator works reasonably well, it falls slightly below the performance of $\infonce$ (about 2\% loss in top-1 accuracy). We noticed the author(s) of \citep{guo2021tight} have updated their empirical estimation procedure since first release of the draft, which we haven't experimented with yet on real data. Also our earlier comparison might not be particular fair as we are comparing single-sample versus multi-sample estimators. So this direction still holds promise, which will be investigated in future work. 

% u-network: slightly lower than simclr

% double networks: explode+ regularize []

{\bf Alternating the updates of $g(x,y)$ and $u(x,y)$.} While our initial attempts with joint optimization of $(g,u)$ failed, we want to use $u$ as a smoothing filtering. This is reminiscent of the exponential moving average trick employed by the $\MINE$, but in a more principled way. Additionally, we further experimented with the idea to optimize on the manifold of $u$ that respects the optimality condition ({\it i.e.}, $u^*(x,y) = g(x,y) + s(x)$, see \citep{guo2021tight} for proofs). Contrary to our expectation, these modification destabilizes training. Our estimators exploded after a few epochs, in a way very similar to the $\DV$ estimator without sufficient negative samples. The exact reason for this is still under investigation. 

\section{Additional Experimental Results}

\subsection{Mini-batch sample MI}
\begin{figure}
  \centering
  \includegraphics[width=.8\textwidth]{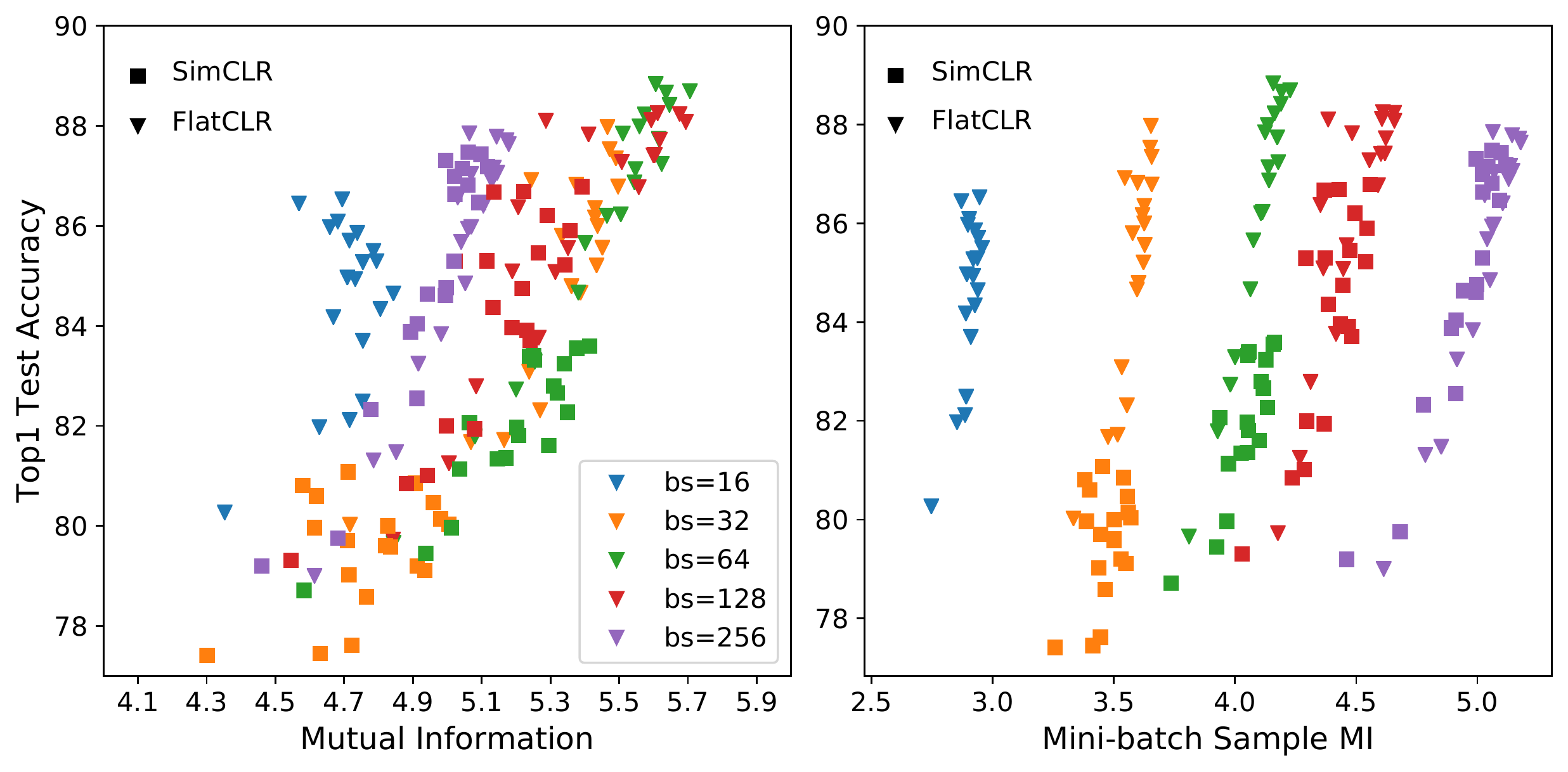}
  \vspace{-.5em}
  \captionof{figure}{While ground-truth representation MI strongly correlates with performance (left), this relation is not evident with the mini-batch sample MI (right). \label{fig:sample_mi_acc} }
\end{figure}

In Figure \ref{fig:sample_mi_acc} we show that mini-batch sample MI is inadequate for predicting downstream performance. 

\subsection{Large-batch training}
\begin{figure}
  \centering
  \includegraphics[width=.5\textwidth]{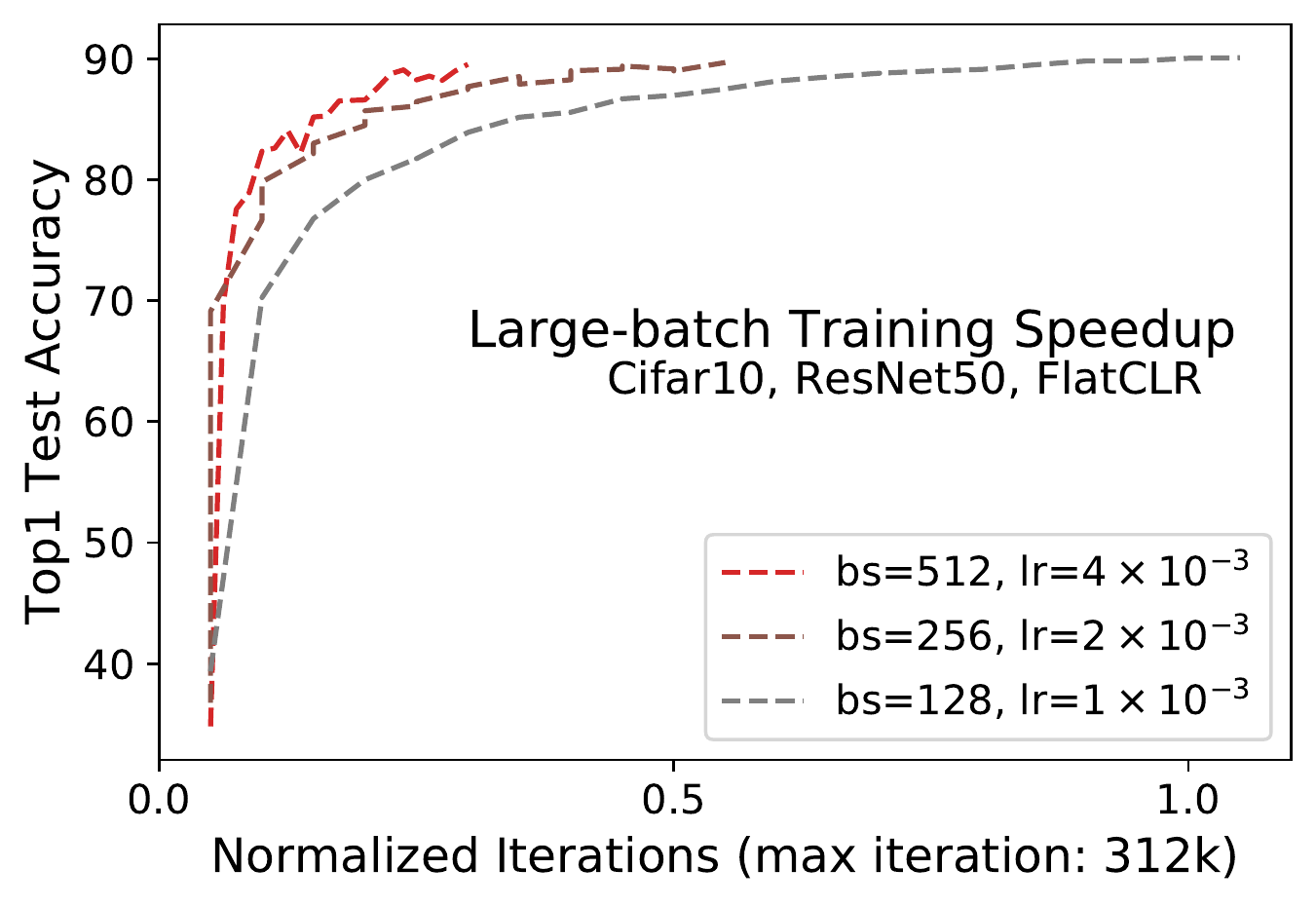}
  \vspace{-1.em}
  \captionof{figure}{Speed up of large-batch training with \texttt{ResNet}-50 on Cifar.  Larger batch leads to faster convergence. \label{fig:speedup_res50} }
\end{figure}

In Figure \ref{fig:speedup_res50} we show large-batch training speedup for the \texttt{ResNet}-50 architecture. Note that we have used the linear scaling of learning rate. And interestingly, for the \texttt{ResNet}-50 architecture model, moderate batch-size (256) actually learned fastest in early training. This implies potential adaptive batch-size strategies to speedup training. 

% \begin{minipage}{\textwidth}
% \end{minipage}

% \begin{minipage}{.355\textwidth}
% \end{minipage}

\subsection{Transfer learning and semi-supervised learning}

\begin{table}[t!]
    \caption{ImageNet SSL transfer learning results.}
    \label{tab:imagenet_trans_supp}
    \centering
    \begin{tabular}{ccccccc}
    \toprule
         Dataset & Cifar10 & Cifar100 & VOC2007 & Flower & SUN397\\
        \midrule
        \multicolumn{6}{c}{\textit{Linear evaluation}}\\
        [2pt]
         $\SimCLR$ & 87.74 & 65.40 & 69.38 & 90.03 & 49.62 \\
         $\FCLR$ & 87.92 & 65.76 & 69.66 & 90.23 & 51.31\\
         [4pt]
        %  \midrule
         \multicolumn{6}{c}{\textit{Fine-tune}}\\
         [2pt]
         $\SimCLR$ & 94.61 & 76.67 & 69.57 & 93.58 & 56.97 \\
         $\FCLR$ & {\bf 95.50} & {\bf 78.92} & {\bf 70.73} & {\bf 95.02} & {\bf 58.37} \\
    \bottomrule
    \end{tabular}
    % \vspace{-1em}
\end{table}

\begin{table}[t!]
\caption{ImageNet SSL results. \label{tab:imagenet_ssl_supp}}
\scalebox{.95}{
\setlength{\tabcolsep}{5pt}
%\begin{center}
\centering    
\begin{tabular}{cccccccccccc}
\toprule
Epoch   & 10    & 20    & 30    & 40    & 50    & 60    & 70    & 80    & 90    & 100   \\
\midrule
\multicolumn{11}{c}{Top-1 Acc} \\
[2pt]
$\SimCLR$ & 40.93 &  46.22 & 48.64 & 50.14 & 52.14 & 53.62 & 55.20 & 56.36 & 56.99 & 57.13 \\ 
$\FCLR$  & {\bf 42.40}  & {\bf 47.69}  & {\bf 49.96}  & {\bf 52.27}  & {\bf 54.11} & {\bf 55.48}  & {\bf 56.98} & {\bf 58.21} & {\bf 58.80}  & {\bf 59.74} \\
[4pt]
\multicolumn{11}{c}{Top-5 Acc} \\
[2pt]
$\SimCLR$  & 65.34 & 70.92 & 73.63 & 75.38 & 76.90 & 78.24& 79.59 & 80.58 & 80.85 & 81.00 \\ 
$\FCLR$  & {\bf 67.17} & {\bf 72.61} & {\bf 74.59} & {\bf 76.77} & {\bf 78.29} & {\bf 79.67} & {\bf 81.06} & {\bf 82.19} & {\bf 82.71} & {\bf 83.18}\\
\bottomrule
\end{tabular}
%\end{center}
}
\end{table}

\begin{table}[t!]
\caption{ImageNet SSL semi-supervised learning results. \label{tab:imagenet_ssl_semi}}
% \scalebox{1.}{
% \setlength{\tabcolsep}{5pt}
\begin{center}
\begin{tabular}{cccccc}
\toprule
Label fraction   & \multicolumn{2}{c}{1\%}  & \multicolumn{2}{c}{10\%}  \\
& Top1 & Top5 & Top1 & Top5 &  \\
\midrule
Supervised &5.25 &14.40  & 41.98 & 67.05\\ 
$\SimCLR$  & 33.44 & 61.29 & 54.62 & 79.89 \\ 
$\FCLR$  & {\bf 36.35} & {\bf 64.59} & {\bf 56.51}  & {\bf 81.32}\\
\bottomrule
\end{tabular}
\end{center}
% }
\end{table}

\paragraph{Transfer Learning via a Linear Classifier}
We trained a logistic regression classifier without $l_2$ regularization on features extracted from the frozen pretrained network. We used Adam to optimize the softmax cross-entropy objective and we did not apply data augmentation. As preprocessing, all images were resized to $224$ pixels along the shorter side using bicubic resampling, after which we took a $224 \times 224$ center crop. 

\paragraph{Transfer Learning via Fine-Tuning} 
We finetuned the entire network using the weights of the pretrained network as initialization. 
We trained for $100$ epochs at a batch size of $512$ using Adam with Nesterov momentum with a momentum parameter of $0.9$. 
At test time, we resized images to $256$ pixels along the shorter side and took a $224 \times 224$ center crop. We fixed the learning rate = $5^{-5}$ and no weight decay in all datasets.
As data augmentation during fine-tuning, we performed only random crops with resize and flips; in contrast to pretraining, we did not perform color augmentation or blurring.

% \citep{chen2020simple} trained for $20,000$ steps at a batch size of $256$ with adjusting batch normalization momentum. They selected the learning rate and weight decay by grid search in every dataset.

\paragraph{Semi-supervised Learning Supervised Baselines} We compare against architecturally identical ResNet models trained on ImageNet with standard cross-entropy loss. These models are trained with the random crops with resize and flip augmentations and are also trained for 100 epochs. 

\subsection{Clarifications on the performance gaps to SOTA results}

This paper aims for promote a novel contrastive learning objective $\FLAT$ that overcomes the limitations of the widely employed $\infonce$. While in all experiment we performed, our $\FLAT$ outperforms $\infonce$ under the same settings, we acknowledge that there is still noticeable performance gap compared to SOTA results reported in literature. We want to emphasize this paper is more about bringing theoretical clarification to the problem, rather than beating SOTA solutions, which requires extensive engineering efforts and significant investment in computation, which we do not possess. 
For example, the $\SimCLR$ paper \citep{chen2020simple} have carried out extensive hyperparameter tuning for each model-dataset combination and select the best hyperparameters on a validation set. The computation resource assessible to us is dwarfed by such need. Their results on transfer learning and semi-supervised learning are transfered from a ResNet50 ($4\times$) (or ResNet50) with $4096$ batch size and $1000$ epochs training on $\SimCLR$.
Our results posted here are transfered from a ResNet50 with $512$ batch size and 100 epochs training on $\SimCLR$ and $\FCLR$. Also, we chose to use the same hyperparameter and training strategy for each dataset to validate the generalization and present a fair comparison between $\SimCLR$ and $\FCLR$.

All in all, the author(s) of this paper is absolutely confident that the proposed $\FCLR$ can help advance SOTA results. We invite the community to achieve this goal together.

\end{document}